\documentclass[lettersize,journal]{IEEEtran}
\usepackage{amsmath,amsfonts}
\usepackage{array}
\usepackage[caption=false,font=normalsize,labelfont=sf,textfont=sf]{subfig}
\usepackage{textcomp}
\usepackage{stfloats}
\usepackage{url}
\usepackage{verbatim}

\usepackage{float}
\usepackage[ruled,linesnumbered]{algorithm2e}
\usepackage{booktabs}
\usepackage[justification=justified,singlelinecheck=false]{caption} 
\usepackage{rotating}
\usepackage{multirow}
\usepackage{caption}

\def\BibTeX{{\rm B\kern-.05em{\sc i\kern-.025em b}\kern-.08em
    T\kern-.1667em\lower.7ex\hbox{E}\kern-.125emX}}
\usepackage{balance}
\begin{document}
\title{Structure-constrained Language-informed Diffusion Model for Unpaired Low-dose Computed Tomography Angiography Reconstruction}
\author{Genyuan Zhang, Zihao Wang, Zhifan Gao, Lei Xu, Zhen Zhou, Haijun Yu, Jianjia Zhang, Xiujian Liu, Weiwei Zhang, Shaoyu Wang, Huazhu Fu, Fenglin Liu, Weiwen Wu
\thanks{G. Zhang, Z. Wang, and F. Liu are with the Key Lab of Optoelectronic Technology and Systems and Engineering Research Center of Industrial Computed Tomography Nondestructive Testing, Ministry of Education, Chongqing University, Chongqing, 400044, China. 
Z. Gao, H. Yu, W. Wu, J. Zhang, X. Liu and W. Zhang are with the School of Biomedical Engineering, Sun Yat-sen University, Shenzhen 518107, China. 
L. Xu and Z. Zhou are with the Department of Radiology, Beijing Anzhen Hospital, Capital Medical University, Beijing 100029, China. 
S. Wang is with the School of Information Engineering, Nanchang, Jiangxi, 330031, China.
H. Fu is with Institute of High Performance Computing (IHPC), Agency for Science, Technology and Research (A*STAR), Singapore 138632, Republic of Singapore. 
The contribution of G. Zhang and Z. Wang are equal. F. Liu and W. Wu are the corresponding authors.}}

\markboth{Journal of \LaTeX\ Class Files,~Vol.~18, No.~9, September~2020}%
{How to Use the IEEEtran \LaTeX \ Templates}

\maketitle

\begin{abstract}
The application of iodinated contrast media (ICM) improves the sensitivity and specificity of computed tomography (CT) for a wide range of clinical indications. 
However, overdose of ICM can cause problems such as kidney damage and life-threatening allergic reactions.
Deep learning methods can generate CT images of normal-dose ICM from low-dose ICM, reducing the required dose while maintaining diagnostic power. 
However, existing methods are difficult to realize accurate enhancement with incompletely paired images, mainly because of the limited ability of the model to recognize specific structures. 
To overcome this limitation, we propose a Structure-constrained Language-informed Diffusion Model (SLDM), a unified medical generation model that integrates structural synergy and spatial intelligence.
First, the structural prior information of the image is effectively extracted to constrain the model inference process, thus ensuring structural consistency in the enhancement process. 
Subsequently, semantic supervision strategy with spatial intelligence is introduced, which integrates the functions of visual perception and spatial reasoning, thus prompting the model to achieve accurate enhancement.
Finally, the subtraction angiography enhancement module is applied, which serves to improve the contrast of the ICM agent region to suitable interval for observation. 
Qualitative analysis of visual comparison and quantitative results of several metrics demonstrate the effectiveness of our method in angiographic reconstruction for low-dose contrast medium CT angiography.
\end{abstract}


\section{Introduction}
\IEEEPARstart{I}{odinated} contrast medium (ICM) is applied in computed tomography angiography (CTA) to enhance tissue contrast for evaluating anatomical structures and pathology \cite{fahling2017understanding,yu2007review}. 
It helps improve the sensitivity and specificity of CT diagnosis for various common clinical indications, such as stroke, trauma, and tumors \cite{chiu2022hypersensitivity}.
However, ICM poses certain risks to patients, including life-threatening allergic reactions such as kidney injury \cite{rubin1997helical,golledge2008acute}. 
To mitigate the patient health risks associated with ICM in CTA, reducing the dose of ICM is a direct approach. 
However, the reduction compromises organ contrast by affecting diagnostic performance\cite{10.1117/12.2581056,kim2021feasibility}. 
Recently, deep learning (DL) algorithms were validated to reduce the contrast dose and synthesize virtual contrast enhanced images \cite{mccollough2015dual,zhang2022multiple,gleeson2004contrast}.

The generative adversarial network (GAN) has been applied to CTA reconstruction\cite{kang2019cycle, lyu2023generative, pang2023ncct}. 
But existing GAN-based CTA enhancement algorithms still face two key challenges including the model's generalization and stability.
First, these models rely on one-shot and end-to-end generation, which enables faster reconstruction speeds, but limits the reliable enforcement of constraints during the reconstruction process \cite{zhu2017unpaired}. 
Second, while GAN-based methods model the target sample distribution through generator-discriminator interactions and learn mapping relationships of image pairs. 
The lack of likelihood evaluation prevents them from fully understanding the physiological features of medical images, thus overlooking the overall organ structure \cite{dhariwal2021diffusion}.
The effectiveness of all these methods is inevitably limited by the inherent flaws of GANs.

\begin{figure}[H]
\centerline{\includegraphics[width=0.45\textwidth]{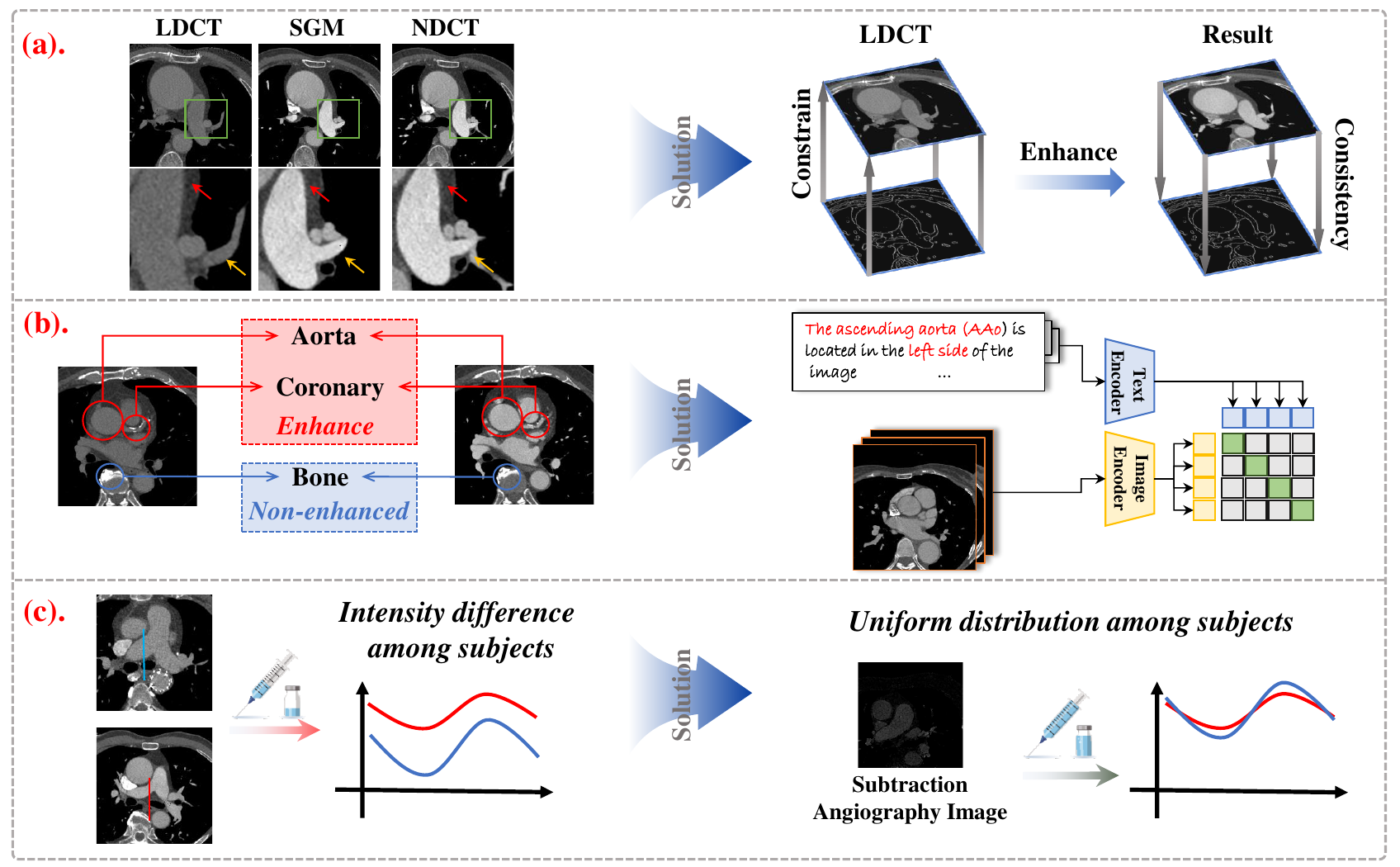}}
\caption{The proposed method employs three strategies to cope with the corresponding problems. (a) Structural differences in low-match data affect reconstructed structures; (b) False enhancements caused by entanglement between enhanced and non-enhanced regions; (c) Intensity differences of subjects affect the reconstructed grayscale.}
\label{motivation}
\end{figure}

The score-based generation model (SGM) adopts explicit likelihood-based representations and a stepwise sampling process to improve sample fidelity in generative modeling tasks\cite{wu2023wavelet,wu2024multi}.
Unfortunately, there are no relevant studies applying SGM to CTA reconstruction.
In-depth analysis, existing SGM-based methods face critical challenges when applied to ICM reconstruction, as shown in Fig. \ref{motivation}.
First, the SGM-based method exhibits false negative reconstruction results, leading to unexpected structural distortions due to mismatches between low-dose CTA (LDCT) images and normal-dose CTA (NDCT) images during acquisition. 
This issue is particularly prominent in small vascular regions and morphological structures, as shown in Fig. 1(a).
Second, the SGM-based method exhibits false positive reconstruction results, where bones that should not be enhanced are incorrectly enhanced, leading to discrepancies with clinical findings, as shown in Fig. 1(b). 
Third, the stability of SGM-based method reconstruction is usually affected by data contrast differences. 
For example, when ICM contrast differences of the aortas are significant in LDCT, the reconstruction results are inconsistent and prone to interference, as shown in Fig. 1(c). 
In summary, addressing the heterogeneous variations induced by ICM is essential for achieving high-quality CTA reconstruction.

To address the above challenges, we propose the Structure-constrained Language-informed Diffusion Model (SLDM) model, which establishes three dimensional collaborative framework of structural prior constraint, semantic text supervision and dynamic contrast enhancement.
First, We propose a structure-constrained mean-reverting stochastic differential equation (SDE). Specifically, our method augments the state space by introducing explicit structural channels. As the intensity components of the SDE diffuse toward potentially misaligned targets, the structural components are constrained to strictly recover to the true topology.
Second, we establish a multi-modal supervision mechanism that encodes anatomical descriptors (e.g., target structure type and spatial coordinates) as text prompts. 
These semantic indicators guide the reconstruction process through cross-modal attention, enhancing spatial awareness in specific clinical contexts. 
Third, the model achieves subtractive contrast enhancement through local contrast adaptation technology. 
Structural and grayscale information from the enhanced regions is precisely extracted to dynamically correct contrast.
As a result, SLDM accomplishes the alignment of specific structures in the complex transformation process.
Our main contributions are as follows.
\begin{itemize}
\item
We introduce a structure-constrained strategy into the image-to-image diffusion model. By using structure channels as anchors, the network is forced to learn conditional identity mappings of geometry, thereby effectively decoupling texture enhancement from structure preservation on unpaired data.
\end{itemize}

\begin{itemize}
\item
Semantic spatial supervision strategy integrating anatomical information is developed. Semantic labels of anatomical structures (such as organ types, locations, and grayscale attributes) are transformed into spatial constraint signals for diffusion models, ensuring that the model achieves precise enhancement of specific structures.
\end{itemize}

\begin{itemize}
\item
The dynamic subtraction angiography enhancement module is designed based on intensity differences. By extracting features of the distribution of contrast agents, locally adaptive contrast guided by dynamic subtraction angiography is used to improve the flexibility of enhancement.
\end{itemize}

\begin{itemize}
\item
We demonstrated efficacy through extensive evaluation on clinical datasets, showing superior reconstruction performance compared to competing methods in preserving both structural fidelity and contrast.
\end{itemize}   

\section{Related Works}
\subsection{Virtual Contrast Enhanced CT Algorithm}
  Most studies have used GAN-based networks to improve the image quality of virtually enhanced contrast CT through adversarial learning \cite{kang2019cycle, lyu2023generative}.
Xie et al. \cite{10.1117/12.2581056} proposed a CycleGAN model to generate contrast-enhanced CT based on simulated non-contrast CT.
Pang et al. \cite{pang2023ncct} proposed two GAN-based synthesizers to perform intersynthesis between LDCT and NDCT images.
Zhu et al. \cite{zhu2017unpaired} proposed a Cycle-GAN framework for recovering contrast-enhanced CT from non-contrast simulated CT by redesigning the cycle consistency loss.
However, the obvious inhomogeneity variation of the ICM against different tissue structures during the enhancement process.
GAN-based methods treat LDCT enhancement as a simple image translation task, so they cannot specifically understand the structural consistency and structural dissimilarity in LDCT enhancement.

\subsection{Visual Language Models Assisted Image Generation}
Currently, significant advances in visual language models (VLM) like contrastive language-image pre-training (CLIP) are revolutionizing medical imaging\cite{dalmaz2022resvit, mokady2021clipcap, rombach2022high}. 
It is conceived that they can effectively integrate comprehensive prior knowledge to provide effective information constraints  \cite{radford2021learning,chambon2022adapting}.
In the medical field, VLMs have been employed for analyzing chest X-ray slices and related radiology reports\cite{niu2023ct}. 
For example, Zhang et al. proposed ConVIRT \cite{zhang2022contrastive}, which learns chest X-ray image representations through bidirectional contrastive learning of image-text pairs, significantly reducing the need for labels in downstream classification tasks.
Chen et al.\cite{chen2024low} proposed a CT denoising method based on linguistically intervened bispace alignment (LEDA) loss, which enhances reconstruction quality by aligning continuous perceptual space with discrete semantic space.
VLMs can be used as generic interfaces for cross-domain image translation \cite{jin2024llmra}, enabling precise domain transformations with task-specific textual cues to synergize multi-source information and specific tasks. 
However, no previous studies have explored the integration of VLMs with the semantic distinction requirements between enhanced regions and non-enhanced regions in CTA reconstruction. 

\subsection{Preliminaries in Image Restoration SDE}

  Image Restoration SDE (IR-SDE) directly simulates the image degradation process via a mean-reverting stochastic differential equation, which effectively circumvents the problem that the degradation process is difficult to accurately estimate \cite{luo2023controlling} \cite{luo2023image}.
During the forward diffusion process, for any state $t \in \left[ {0,T} \right]$, the IR-SDE is redefined as a conditional Ornstein-Uhlenbeck process\cite{yue2023image}:
\begin{equation}
\label{eqn-1}
d{\bf{x}} = {\theta _t}\left( {\mu - {\bf{x}}} \right)dt + {\sigma _t}d\omega,
\end{equation}
where $\mu$ is is the state mean, and $\theta_t$ and $\sigma_t$ are positive parameters that vary with time $t$, and they describe the mean-reversion speed and stochastic volatility in the diffusion process, respectively.
$\omega$ is a standard Wiener process\cite{arjovsky2017towards}, which brings randomness to the differential equation. 

The reverse-time representation is given by:
\begin{equation}
\begin{array}{l}
d\mathbf{x} = \theta_t \left( \mu - \mathbf{x} \right) - \sigma_t^2 \nabla_{\mathbf{x}} \log p_t(\mathbf{x}) + \sigma_t d\widehat{\omega},
\end{array}
\end{equation}
where $\widehat \omega$ is a reverse-time Wiener process, $\nabla _{{\bf{x}}}\log {p_t}({\bf{x}})$ is the score function, which can be approximated by training a time-dependent neural network $s_\phi$.

In order to obtain exact scores, we rigorously derive the optimal inverse state during the diffusion process. 
The optimal inversion state $\bf{x}_{t - 1}^*$ at step $t-1$ can be naturally obtained from $\bf{x}_{t}$ via the most negative log-likelihood, while this process is performed under the constraints of the topology $\bf{y}_{t}$:
\begin{equation}
  {\bf{x}}_{t - 1}^* = \operatorname*{argmin}_{{{\bf{x}}_{t - 1}}} [ - \log q({{\bf{x}}_{t - 1}}|{{\bf{x}}_t},{{\bf{x}}_{0}})],
\end{equation}
where ${\bf{x}}_{t - 1}^*$ denotes the ideal state obtained in the reverse direction from ${\bf{x}}_{t}$. 
  
By solving the above objective, the score prediction network is trained with maximum likelihood loss, which specifies the optimal backward path $\bf{x}_{t - 1}^*$ for all times\cite{liu2024structure}:
\begin{equation}\label{eqn-5}
\begin{array}{l}
{\bf{x}}_{t - 1}^* = \frac{{1 - {e^{ - 2{{\bar \theta }_{t - 1}}}}}}{{1 - {e^{ - 2{{\bar \theta }_t}}}}}{e^{\theta '_t}}\left( {{{\bf{x}}_t} - {\mu _{\bf{}}}} \right)\\
 + \frac{{1 - {e^{ - 2{{\theta '_t}}}}}}{{1 - {e^{ - 2{{\bar \theta }_t}}}}}{e^{ - {{\bar \theta }_{t - 1}}}}\left( {{{\bf{x}}_0} - {\mu _{\bf{}}}} \right) + {\mu},
\end{array}
\end{equation}
where ${\theta'_t} = \int_{t-1}^t {{\theta _t}dt}$ and ${\bar \theta _t} = \int_0^t {{\theta_t}dt} $. 
  Based on Eq. (4), $\bf{x}_{t - 1}^*$ is obtained using ${{\bf{x}}_t}$ during diffusion.
Hence, IRSDE can optimize $\nabla _{{\bf{x}}}\log {p_t}({\bf{x}})$ through the following training target:
\begin{equation}\label{eqn-6}
\begin{array}{l}
{L_\gamma }\left( \phi  \right) = \sum\nolimits_{t = 1}^T {{\gamma _t}{\rm{E}}\left[ {\left\| {{\nabla _{\bf{x}}}\log {p_t}({\bf{x}}) - {s_\phi }({{\bf{x}}_t},{\bf{\mu }},t)} \right\|} \right]},
\end{array}
\end{equation}
where ${{\gamma _t}}$ represents the weight at time $t$, and ${{s_\phi }({{\bf{x}}_t},{\bf{\mu }},t)}$ denotes the score of state ${\bf{x}}$ under condition $\mu$ at time $t$. 
After training, IRSDE can utilize ${{s_\phi }({{\bf{x}}_t},{\bf{\mu }},t)}$ to iteratively generate high-quality images via a numerical solver, starting from a noisy initial state ${\bf{x}_T}$.

\begin{figure*}
\centerline{\includegraphics[width=\textwidth]{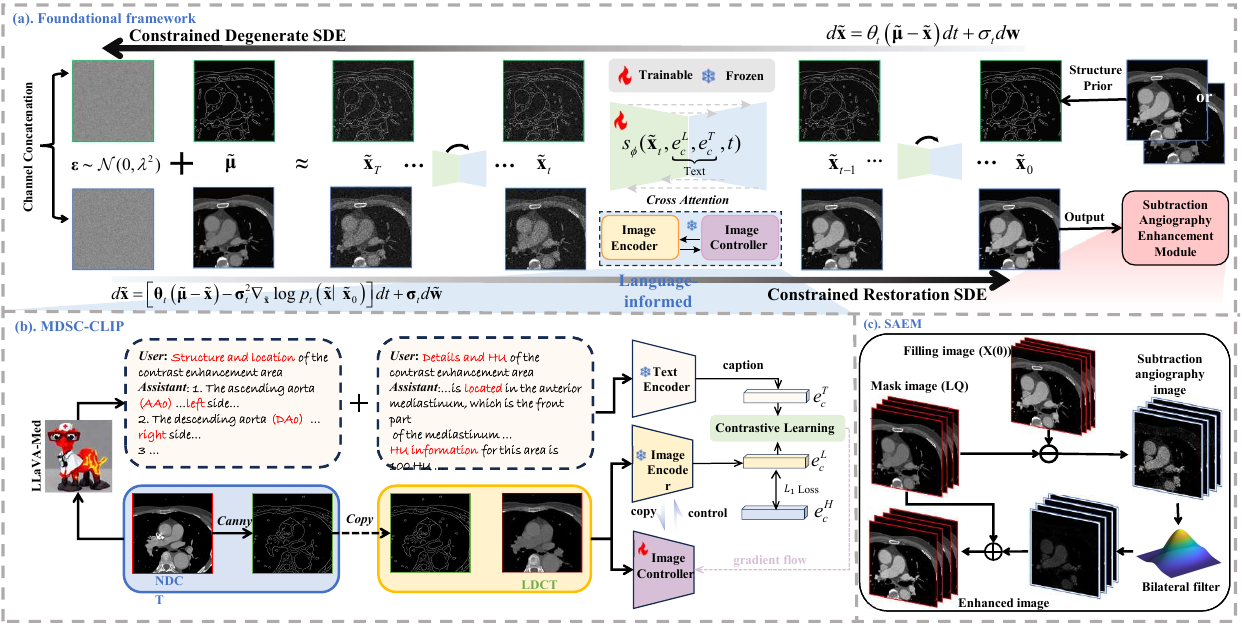}}
\caption{Overview of the proposed SLDM. (a) The IR-SDE constitutes the foundational framework, topological constraint score (green) and text supervision score (blue) are integrated throughout both training and testing phases to ensure structural fidelity; (b) The pre-trained CTA-CLIP encoder, designed to facilitate fine-grained feature extraction and semantic alignment within the model; (c) SAEM refines the reconstruction process through targeted optimization of subtraction angiography features.}
\label{fig1}
\end{figure*}

\vspace{-0.3cm}
\section{METHODS}
\subsection{The Overall of Problem and Method}
\subsubsection{Overview of the ICM Enhancement Process}
  The core challenge in low-dose CTA reconstruction is generating NDCT images that meet the following criteria:
(1) Structural consistency: NDCT image must accurately preserve anatomical structures of LDCT image(e.g., vascular branches, bone positions); 
(2) Contrast accuracy: Vascular regions in NDCT image are enhanced to an appropriate HU range while preventing false enhancement in non-vascular areas such as bones;
(3) Clinical adaptability: NDCT image must be adjustable to meet individualized diagnostic needs, such as contrast tuning for different vessel sizes.
Therefore, the enhancement process has an obvious disentangled representation property \cite{gonzalez2018image}. 
Virtual enhancement algorithms should fully recognize this problem. 
For the disentangled representation method, it can be analyzed from the perspective of information decoupling. 

  We analyze the key challenge in terms of global information transformation and local information transformation.
From the perspective of global information, we can regard the domain conversion from LDCT image to NDCT image.
LDCT image and NDCT image can be considered as originating from different the data distribution, respectively.
However, this transformation is complex and difficult to estimate accurately because an exact mathematic representation of the underlying transformation process is unavailable.
From the perspective of local information, the enhancement process can be modeled as involving structural consistency (e.g., bones) and structural dissimilarity (e.g., coronary artery). 
Therefore, the core of achieving high-quality enhancement lies in achieving local structural enhancement of ICM-induced regions and precise preservation of other regions.
Our SLDM addresses this limitation through precise structural constraints and specific structural prompts to ensure better synthesis results.

\subsubsection{Architecture of SLDM}
The overall workflow of SLDM is depicted in Fig. 2 and consists of three main phases: the diffusion model prior generation phase, the CTA-CLIP assistance phase, and the subtraction angiography enhancement phase. 
In the prior generation phase (Fig. 2(a)), topological constraint is introduced to extract prior topological spatial information to maintain precise structural consistency during the enhancement process.
In the CTA-CLIP assistance phase (Fig. 2(b)), semantic supervision is introduced to enable structure perception of different anatomical structures during the enhancement process.
CTA-CLIP fine-tunes the reverse diffusion process under the guidance of the cross-attention mechanism.
It accurately enhances contrast areas by utilizing text prompts of specific structures.
In the subtraction angiography enhancement phase (Fig. 2(c)), the structural information of the contrast agent regions is accurately extracted in the subtraction angiography enhancement module (SAEM). 
These regions are dynamically optimized to achieve more flexible and adaptive reconstruction results.

\subsubsection{Overall Clarification}
The training process consists of two stages: fine-tuning of CTA-CLIP and training of the topologically-constrained IR-SDE.
The core of CTA-CLIP is to use contrastive learning to align the semantic representations of NDCT images with the image features of LDCT.
Topologically-constrained IR-SDE is trained using LDCT-NDCT data pairs, while the topological space and text information of NDCT are used as constraints. 
During the testing process, LDCT images are gradually enhanced by leveraging the text-guided and topology-guided score functions learned during training and the optimal reverse-time path.
Finally, SAEM further refines the reconstructed NDCT images.
The overall SLDM for the training and testing process is outlined in Algorithm 1.

\begin{algorithm}
    \SetAlgoLined 
    \caption{SLDM: Training and Sampling Process} 
    \KwData{LDCT data, NDCT data} 
    \KwIn{LDCT image: ${\bf{y}}$, NDCT image: ${{\bf{x}}}$, Topology: $\bf{s}$, parameters: ${{\bf{\sigma }}_{t }}$, ${({{\theta _t}})_{t \in [1,T]}}$, $T$, $\lambda$} 
    \KwOut{Enhanced image: ${{\bf{x}}_{out}}$} 
    
    \hspace*{\fill} \textbf{\uppercase\expandafter{\romannumeral1}. Training Process} \hspace*{\fill} \\
    
    \hspace*{\fill} \textit{A: CTA-CLIP Model Pre-training} \hspace*{\fill} \\
    
    Get text description by LLaVA-Med: (NDCT data) $\to c_{Text}$ \;
    Generate text and image embedding: $\{ e_c^T,e_c^L\}$ \;
    Use Eq. (7) to calculate the contrastive learning loss \;
    
    \hspace*{\fill} \textit{B: Topology-constrained IR-SDE Training} \hspace*{\fill} \\
    
    Obtain score function: ${S_\phi }$ // The score obtained by combining topological constraint and textual constraint \;
    
    \hspace*{\fill} \textbf{\uppercase\expandafter{\romannumeral2}. Testing Process} \hspace*{\fill} \\
    
    Data initialization: ${\bf{\tilde x}}(t) = \left[ {{\bf{x}}(t),{\bf{s}}(t)} \right]$\;
    \For{$t = T$ \textbf{down to} $0$}{
        $\theta '_i: = \int_{i - 1}^i {{\theta _t}dt},\; {{\bar \theta }_t} = \int_0^t {{\theta _j}} dj,\; \sigma _t^2/{\theta _t} = 2{\kappa ^2},\; {v_t} = {\kappa ^2}(1 - {e^{ - 2{{\bar \theta }_t}}})$ // Calculate required parameters \;
        ${\bf{\hat x}}_0 = {e^{{{\bar \theta }_t}}}({{\bf{\tilde x}}_t} - \mu _t - \sqrt {{v_t}} {{\bf{s}}_\phi}) + \mu _t$ // Estimate initial value \; 
        ${\bf{\tilde x}}_{t - 1}^* = \frac{{1 - {e^{ - 2{{\bar \theta }_{t - 1}}}}}}{{1 - {e^{ - 2{{\bar \theta }_t}}}}}{e^{\theta '_t}}\left( {{{\bf{\tilde x}}_t} - {\mu }} \right) + {\mu } + \frac{{1 - {e^{ - 2{{\theta '_t}}}}}}{{1 - {e^{ - 2{{\bar \theta }_t}}}}}{e^{ - {{\bar \theta }_{t - 1}}}}\left( {{\bf{\hat x}}_0 - {\mu }} \right)$ \;
        ${\tilde \beta }_t = \frac{{(1 - {e^{ - 2{{\bar \theta }_{t - 1}}}})(1 - {e^{ - 2{{\theta '}_t}}})}}{{1 - {e^{ - 2{{\bar \theta }_t}}}}}$\;
        $p({\bf{\tilde x} _{t-1}}|\bf{\tilde x} _t,{\bf{\hat x}}_0) = {\cal N}(\bf{\tilde x} _{t-1}|{\bf{\tilde x}}_{t - 1}^*,{{\tilde \beta }_t}{\bf{I}})$ // Posterior distribution; 
    }
    
    \hspace*{\fill} \textit{C: Subtraction angiography enhancement} \hspace*{\fill} \\
    
    ${{{\bf{x}}}_{fill}} = {{{\bf{x}}}_{0}}, {{{\bf{x}}}_{mask}} = \bf{y}$  \; 
    Compute subtraction: $\bf{x}_{sub} = {{{\bf{x}}}_{mask}} - {{\bf{x}}_{fill}}$ \;
    Apply filter: $\bf{x}_{sub}^B = \text{Bilateral}({{\bf{x}}_{sub}})$ \;
    Final output: $\bf{x}_{out} = {{\bf{x}}_{mask}} + \lambda {\bf{x}}_{sub}^B$ \;
    \textbf{return} $\bf{x}_{out}$
\end{algorithm}

\subsection{Topologically Constrained IR-SDE}  

Although NDCT and LDCT obtained in clinical settings may come from the same patient and the same equipment, patient displacement or respiration can cause mismatches between NDCT and LDCT. This results in significant structural distortions in the image-to-image diffusion model \cite{luo2023image}\cite{liu20232}\cite{shi2024resfusion}. We develop our theory based on a IR-SDE. 
We consider two image domains: a NDCT domain ${\bf{x}} \in \mathcal{X}$  and a LDCT domain ${\bf{y}} \in\mathcal{Y}$. The images in these two domains are weakly paired. Our goal is to learn a restoration function that maps the LDCT to NDCT while preserving the structure of the LDCT. We introduce a structure extraction function $C$, where $C(\bf{x})$ represents the structural information of any image $\bf{x}$. During the training phase, we construct extended training pairs:
\begin{equation}
\label{e1}
{{{\bf{\tilde x}}}_0} = \left[ {{{\bf{x}}_0},C({{\bf{x}}_0})} \right],\mathbf {\tilde {\mu}} = \left[ {{{\bf{y}}_0},C({{\bf{x}}_0})} \right].
\end{equation}

  We consider an extended state ${\bf{\tilde x}}(t) = \left[ {{\bf{x}}(t),{\bf{s}}(t)} \right]$, where ${\bf{x}}(t)$ is image part, and ${\bf{s}}(t)$ is structure part. Therefore, we propose the forward and backward processes of topological constraint IR-SDE as follows:
\begin{equation}
\label{e2}
\left\{ \begin{array}{l}
d{\bf{\tilde x}} = {\theta _t}\left( {{\bf{\tilde \mu }} - {\bf{\tilde x}}} \right)dt + {\sigma _t}d{{\omega}}\\
d{\bf{\tilde x}} = \left[ {{{\bf{\theta }}_t}\left( {{\bf{\tilde \mu }} - {\bf{\tilde x}}} \right) - {\bf{\sigma }}_t^2{\nabla _{{\bf{\tilde x}}}}\log {p_t}\left( {{\bf{\tilde x}}} \right)} \right]dt + {{\bf{\sigma }}_t}d{\bf{\tilde{\omega}}}
\end{array} \right.
\end{equation}

  Under the maximum likelihood learning (   ), we optimize the network to match the optimal reverse solution (   ). For the structural channel, the optimal reverse solution is:
\begin{equation}
\label{e3}
\begin{array}{l}
{\bf{s}}_{i - 1}^* = \frac{{1 - {e^{ - {{\tilde \theta }_{i - 1}}}}}}{{1 - {e^{ - 2 \tilde \theta_i}}}}{e^{ - {{\theta '}_i}}}({{\bf{s}}_i} - {{\bf{s}}_{{\rm{gt}}}}) + \\
\frac{{1 - {e^{ - 2{{\theta '}_{i}}}}}}{{1 - {e^{ - 2{{\tilde \theta }_i}}}}}{e^{ - {{\tilde \theta }_{i - 1}}}}({{\bf{s}}_{{\rm{gt}}}} - {{\bf{s}}_{{\rm{gt}}}}) + {{\bf{s}}_{{\rm{gt}}}}
\end{array},
\end{equation}
where ${{\bf{s}}_{{\rm{gt}}}} = C({{\bf{x}}_0})$.
If $\bf{s}_i = \bf{s}_{\text{gt}}$, then $\bf{s}_{i-1}^* = \textbf{s}_{\text{gt}}$. Therefore, during training, when the initial mean of ${\bf{s}}(T)$ is ${\bf{s}}_{\text{gt}}$, the network learns to maintain the structural channels as $\bf{s}_{\text{gt}}$. 

During the testing phase, the input to the structural channels is changed to ${\bf{s}_{\text{lq}}}={C}({{\bf{y}}_0})$. \\
\noindent \textbf{Theorem} (Consistency of generated results): Let the image generated during the testing phase be $\hat{\bf{x}}_0$, and the corresponding structural component be ${\bf{s}}_0$. Then, when the backward step number $T \to \infty$:\\
\noindent 1. $\mathop {\lim }\limits_{T \to \infty } \mathbb {E}\left[ {{{\left\| {{{\bf{s}}_0} - {\bf{s}_{{\rm{lq}}}}} \right\|}_2}} \right] = 0$.\\
\noindent 2. There exists a constant $K > 0$ such that:
\begin{equation}
\label{e5}
\mathbb {E}\left[ {{{\left\| {C({{{\bf{\tilde x}}}_0}) - {{\bf{s}}_{\rm{lq}}}} \right\|}_2}} \right] \le K \mathbb {E}\left[ {{{\left\| {{{\bf{s}}_0} - {{\bf{s}}_{{\rm{lq}}}}} \right\|}_2}} \right]
\end{equation}
The proof is provided in Appendix A. The above theorem shows that the structure of the generated result is constrained by the structure of the test input. Ultimately, our design extends IR-SDE to conditions where weakly paired data can be used, enhancing LDCT while preserving the LDCT structure.

\subsection{CTA-CLIP Assistance Phase}
To achieve precise enhancement of specific anatomical structures, we introduce CTA-CLIP with spatial intelligence integrating visual perception and spatial reasoning capabilities.
The core idea is to build a dedicated CLIP variant, termed CTA-CLIP, specifically for the CTA enhancement task.
Subsequently, CTA-CLIP is embedded into the reverse process of the diffusion model to achieve dynamic guidance and optimization.
The key to visual perception lies in constructing a prompt learning framework, which establishes structure-aware mapping from image to text.
Meanwhile, spatial reasoning is realized by dynamically guiding the generation process via the cross-attention module.
In the following, we elaborate on the mathematic mechanism of semantic-guided precise enhancement.

First, we pre-train CTA-CLIP using effective text and image information.
The core components of CTA-CLIP include a text encoder (Transformer) and an image encoder (Vision Transformer), which embed LDCT images to match the corresponding text features of NDCT images\cite{radford2021learning}.
For text embedding, we employ LLAVA-med which is a medical large language model (LLM) developed by Microsoft to generate corresponding text description features \cite{li2023llava}.
Specifically, LLAVA-med analyzes the structure of NDCT images and obtains specific feature descriptions, including organ types, locations, and regions requiring enhancement. 
The text encoder subsequently encodes task-specific information into text embeddings $e_c^L$.
Visual information is encoded into embeddings via the image embeddings $e_c^T$.

  To align text and image embeddings, we calculate the cosine similarity between the embedding pairs though cross-entropy loss \cite{luo2023controlling}.
The loss can be simplified as follows:
\begin{equation}
\setlength{\abovedisplayskip}{1pt}
\setlength{\belowdisplayskip}{1pt}
{\cal L}\left( {a,b} \right) =  - \log \left( {\frac{{{e^{\cos (a,{b^ + })}}}}{{\sum {{e^{\cos (a,{b^ * })}}} }}} \right),
\end{equation}
where $a$ represents LDCT image, $b^+$ represents the text of the matched positive sample, $b^*$ represents all texts (including texts of unmatched negative samples), $\cos$ denoting cosine similarity.
The core of contrastive learning is to maximize the probability of the similarity of positive samples among the similarities of all sample pairs.
By maximizing the similarity probability of positive sample pairs, CTA-CLIP can achieve cross-modal image-semantic alignment.

After completing the pre-training of CTA-CLIP, a cross-attention module is introduced to integrate the text descriptions and prompt functions into the reverse enhancement process\cite{niu2023ct}.
The output of the text encoder of feature descriptions can guide the diffusion model to generate images that conform to the text descriptions.
The underlying mechanism lies in transforming the diffusion model into a more flexible conditional image generator.
To preprocess textual cues from linguistic cues ${c_{Text}}$, a structure-specific encoder is introduced that projects ${c_{Text}}$ to an intermediate representation, which is then passed through the implementation of 
\begin{equation}
Attention\left( {Q,K,V} \right) = softmax \left( {\frac{{Q{K^T}}}{{\sqrt d }}} \right) \cdot V, 
\end{equation}
where $Q$ (Query), $K$ (Key), $V$ (Value) are learnable projection matrices, $d$ is the task token dimension.
In the cross-attention mechanism, $Q$ is generated from image features in the U-Net, while $K$ and $V$ are generated from semantic vectors of the text encoder. Through the cross-attention mechanism, the image features in the diffusion model enhancement process will be aligned with the corresponding text information, thereby being guided by the text information during the generation process.

\subsection{Subtraction Angiography Enhancement Module}
  Inspired by digital subtraction angiography (DSA) technology, we designed the SAEM algorithm based on time-domain subtraction to adapt to individual vascular characteristics.
SAEM designates LDCT as the mask image ${{\bf{x}}_{mask}}$, while considering the enhanced image ${{\bf{x}}_{0}}$ as the filling image ${{\bf{x}}_{fill}}$. The reconstruction process can be viewed as the process of increasing the contrast agent dose through virtual process.
Then, the precise structure of the enhanced region is obtained and optimized to obtain better quality vascular imaging.
Dynamic adjustment of vessel grayscale information can be achieved by subtraction images.

  As depicted in Fig. 2(c), SAEM acquires the subtracted image by computing the disparity between the mask image and the filling image. 
In the subtracted image, the background manifestations are eradicated, leaving only the vascular images that contain the contrast agent. 

  First, rough subtraction angiography image ${{\bf{x}}_{sub}}$ is obtained from the mask image ${{\bf{x}}_{mask}}$ and the filling image ${{\bf{x}}_{fill}}$:
\begin{equation}{{\bf{x}}_{fill}} - {{\bf{x}}_{mask}} = {{\bf{x}}_{sub}}.\end{equation}

  Then, the rough subtraction angiography image is denoised to obtain structurally smooth subtraction angiography image ${\bf{x}}_{_{res}}^B$, which represents the effective contrast's region:
\begin{equation}
{\bf{x}}_{_{sub}}^B = {\bf{Bilateral}}\left( {{{\bf{x}}_{sub}}} \right).
\label{eq}
\end{equation}

Finally, the smooth subtraction angiography image is fused with LDCT by weighting parameter $\lambda $, which obtains accurate reconstructed image ${{\bf{x}}_{out}}$ and realizes dose's adjustment:
\begin{equation}
{{\bf{x}}_{result}} = {{\bf{x}}_{mask}} + \lambda {\bf{x}}_{_{sub}}^B.
\label{eq2}
\end{equation}

Thus, SAEM confers on SLDM the superiority of the ability to achieve controllable grayscale adjustment targeting the enhancement region.
The benefit of this simple module is obvious and meaningful. 
It allows physician the flexibility to customize vascular enhancements rather than the traditional controlled-dose regimen based on the patient’s pathology.

\section{EXPERIMENTS }
\subsection{Datasets}
\subsubsection{CTA Dataset Acquisition and Pre-processing}
Datasets were acquired from aortography-enhanced CTA scans at Beijing Anzhen Hospital of Capital Medical University.
For each subject, both NDCT and LDCT scans were performed, covering the anatomical range from the lung apices to the lesser trochanter of the femur.
All clinical tests were approved by the Medical Ethics Committee of Beijing Anzhen Hospital (Ethics approval number: 2022 Kelun Review No. (26), approval date: December 28, 2022).
NDCT examinations were carried out on a 256-slice multi-detector system (Revolution CT, GE Healthcare, United States).
Iodixanol 320 (320 mg I/ml) was administered intravenously at a volume of 40–70 ml.
Tube potential was tailored to patient BMI, ranging from 80 to 120 kV, while tube current was automatically adjusted with CareDose 4D technology. 
The acquisition yielded an in-plane spatial resolution of 0.3–0.5 mm and a nominal slice thickness of 0.625 mm.
LDCT scans used the same volume of a 1:2 mixture of iodixanol 320 and saline, with scanning parameters matching those of the NDCT protocol except for the iodinated contrast dose.
The dataset comprises normal-dose and low-dose CTA scans from 50 patients, 40 patients were used for training, 5 patients for verification and 5 patients for testing. 
Each patient provided 224 CT slices for the study.
All subsets include images covering critical regions (aorta, coronary arteries, pulmonary veins), avoiding bias toward specific body parts.
Hierarchical segmentation was not employed to preserve patient-specific anatomical continuity.
It should be noted that due to the limitations of multiple acquisitions, the NDCT is not ground truth image of the LDCT.
Therefore, since the LDCT-NDCT data pairs are not perfectly matched, NDCT can only serve as a reference image for LDCT.

  To enhance visualization of arterial-tissue intensity differences, windowing adjustments were applied to both LDCT and NDCT datasets. For LDCT, intensities ranged from -1024 HU to 3071 HU, with the aorta at -120 HU and adjacent vessels at -60 HU. 
NDCT showed a theoretical maximum arterial enhancement of 700 HU. A window level of 200 HU and width of 1000 HU were empirically set, with intensities outside this range truncated: values below -300 HU were reset to -300 HU, and those above 700 HU to 700 HU. 
Both LDCT and NDCT datasets underwent identical truncation and were normalized to a scale of 0-1 for network input. 
Additionally, Gaussian denoising was applied to optimize image quality, effectively reducing noise while preserving structural information.
\subsubsection{Preparation of Topological Structure Dataset}
A carefully designed topological structure dataset was constructed to extract structural information. 
A three-channel data composition approach, structured as Gray-Topology-Topology, was employed.
The canny operator was used to generate topological images. 
For the train set, the first channel of each input pair used grayscale images from both NDCT and LDCT, while the remaining channels utilized topological structure of NDCT.
This method prevents structural deformation during domain transformation.
For the test set, the first channel relied solely on LDCT grayscale images, with the remaining channels constructed from topological structure of LDCT. 
The structural constraints of the enhancement process can be realized through this production of datasets.

\subsubsection{Natural Language Dataset Construction}
The semantic supervision datasets were constructed using NDCT images, and corresponding text descriptions(Fig. 3(a)). 
Text descriptions are obtained through LLAVA-med-v1.5-mistral-7b which is a medical LLM developed by Microsoft \cite{li2023llava}.
Owing to the emergent capacity limitation of LLM, they cannot accurately respond to questions when multiple cue words were input simultaneously. 
To improve the model's ability to understand question details, we employ a multi-semantic prompt strategy. 
In details, we designed two prompts leveraging the LLM's contextual information processing capabilities. 
The first prompt requested a detailed description of structures and their locations in the contrast-enhanced CT section. 
The second prompt asked for detailed information on the location and Hounsfield units of the contrast-enhanced area. 
A text file was generated from these prompt words, offering a detailed summary of structural information on anatomical structures such as the aorta, coronary artery, types of organs and bones, and the corresponding locations in the NDCT images.
Additionally, it provided precise details about the location of the contrast agent regions and the gray scale values (as shown in Fig. 3(b)).
Next, image–text embeddings were produced by leveraging the pre-trained encoders within CTA-CLIP.
Specifically, CTA-CLIP’s image and text encoders adopt ViT-B-32, a Vision Transformer variant, as their shared backbone.

\begin{figure}[!ht]
\centerline{\includegraphics[width=0.5\textwidth]{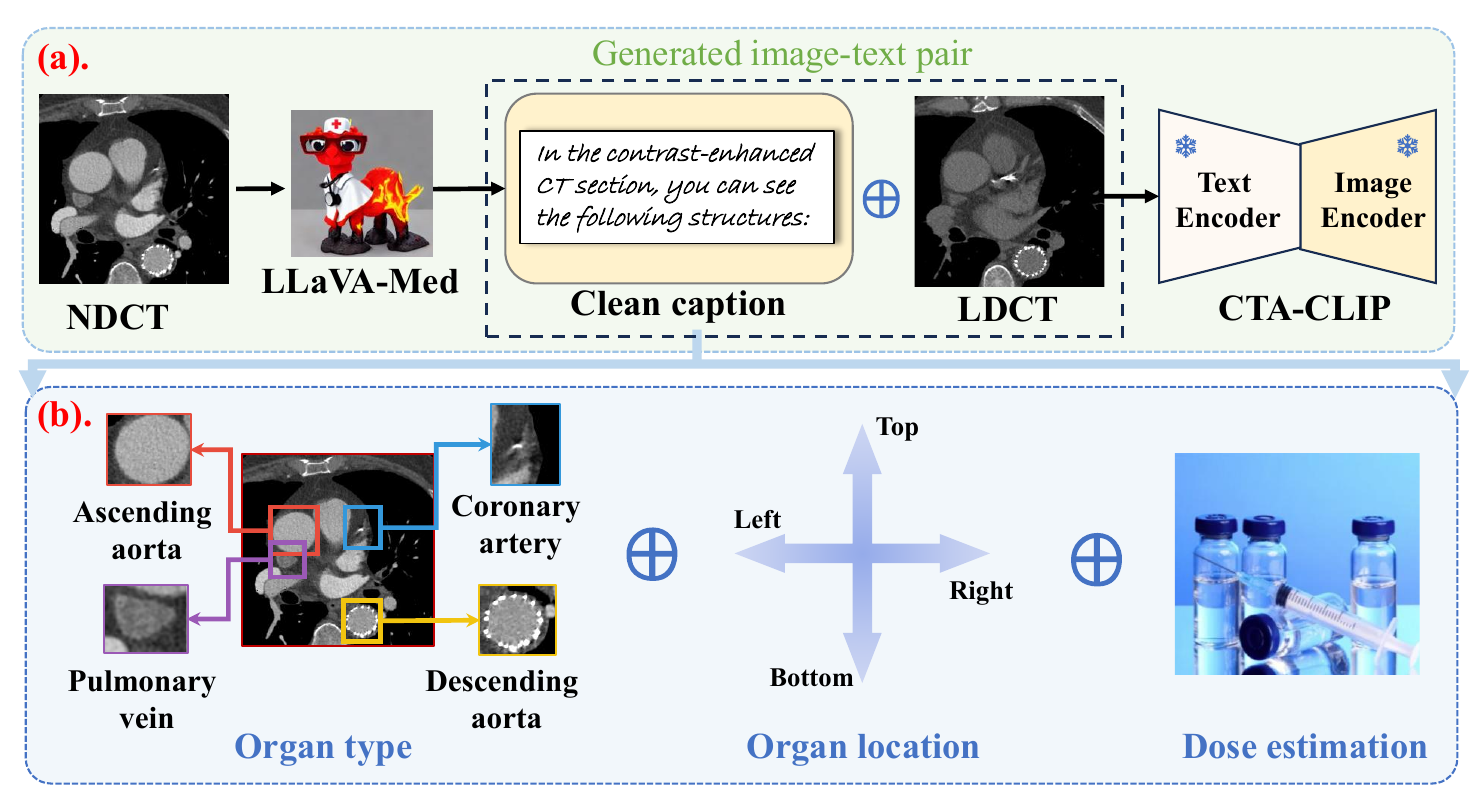}}
\caption{Validation semantic datasets of the proposed SLDM. (a) An example of generating the LDCT image-text tuple via LLAVA-med; (b) The image description text of NDCT included information on the type of organ, location information and the HU value of contrast agent area.}
\label{fig2}
\end{figure}

\subsection{Evaluation Metrics}
The evaluation involves the state-of-the-art image translation methods, including MUNIT \cite{huang2018multimodal}, IdentityGAN \cite{kang2019cycle}, Cycle-GAN \cite{zhu2017unpaired}, and MALAR \cite{zhang2022multiple}. These studies have been widely applied in the field of CTA reconstruction.
Due to the current lack of diffusion model-based CTA reconstruction algorithms, we applied two diffusion mod    ·el algorithms designed for image translation tasks as the control group for this study.
Syn-Diffusion \cite{ozbey2023unsupervised} and FGDM \cite{10287612}, medical image translation framework derived from diffusion model, were modified and applied to LDCT reconstruction.

To quantitatively assess and compare the validity of the reconstruction results, peak signal-to-noise ratio (PSNR) and structural similarity (SSIM) metrics were calculated for the similar region of interest (ROI) region of the result and NDCT, and an increase in PSNR and SSIM values corresponded to an improvement in reconstruction quality. 
In addition, two reference-free image metrics were used.
Signal-to-noise ratio (SNR) measures the ratio of signal intensity (vascular region) to noise intensity (background region). 
A higher SNR value indicates clearer vascular signals.
The improved signal-to-noise ratio (ISNR) quantifies the degree of SNR enhancement relative to LDCT.
If ISNR\textgreater0, the larger the ISNR, the greater the improvement relative to degraded image restoration and the better the algorithm's image restoration capability.

To investigate the generalizability of various reconstruction methods to clinically realistic LDCT images, we randomly selected reconstructed images for subjective evaluation. Two experienced radiologists performed blinded quantitative evaluation of results. 
Evaluation criteria included image sharpness, noise level, contrast, and detail preservation. We used an integer score range from 1 to 5, with higher scores indicating better quality. 
Specifically, a score of 5 indicates good contrast, good vessel circling, few artifacts, and easy diagnosis; a score of 1 indicates unexplained contrast, severe artifacts, indistinguishable vessel circling, and no diagnosis. 
All subjects were anonymous and randomly shuffled. Two radiographers had no knowledge of the scanning and processing conditions to minimize environmental bias. 
We calculated the mean score and corresponding confidence intervals for each radiologist.

\begin{table}
\centering
\caption{QUANTITATIVE EVALUATION (MEAN ± STD) FROM SELECTED STRUCTURES}
\resizebox{1\columnwidth}{!}{
\begin{tabular}{ccccc} 
\toprule
   Method                 & SNR   & ISNR   & PSNR    & SSIM  \\ 
\midrule
MUNIT   & 15.44±2.48  & 6.92±2.45    & 19.04±2.47    & 0.7545±0.0024    \\
IdentityGAN  & 20.74±4.34      & 12.22±2.85    &  24.31±4.71     &  0.8023±0.0010 \\
Cycle-GAN  & 20.42±2.92    & 11.90±4.05     & 23.98±3.05    & 0.7980±0.0012    \\
MALAR  & 19.18±1.09  & 10.70±7.10 & 22.74±1.60 & 0.7940±0.0012    \\
Syn-diffusion & 16.60±7.38 & 8.08±4.38 & 20.17±7.78 & 0.7847±0.0015  \\
FGDM & 19.99±1.15 & 10.92±1.05 & 23.55±1.19 & 0.7816±0.0008  \\
\pmb{SLDM} &  \pmb{20.87±2.42}  &  \pmb{12.35±3.51} &  \pmb{24.44±2.58}  &  \pmb{0.8050±0.0015} \\
\bottomrule
\end{tabular}
}
\label{table 1}
\end{table}


\subsection{Training Details}
Spatial topological structure is used as conditions in the training process. The forward and backward diffusion processes of the model were supervised by the spatial topological structure.  We used the Adam optimizer with $\beta_1$ = 0.9, $\beta_2$ = 0.99. 
The total training steps are fixed to 700 thousand and the initial learning rate set to $10^{-4}$ and decays half per 200 thousand iterations. The noise level $\sigma$ and the number of diffusion denoising steps $T$ was fixed to 50 and 100.The stationary variance $\lambda^2$ is set to 10. The batch size was set to 16. Our model was implemented using PyTorch and trained about 500,000 iterations on RTX 4090 GPU.
\begin{figure}[!ht]
\centerline{\includegraphics[width=0.46\textwidth]{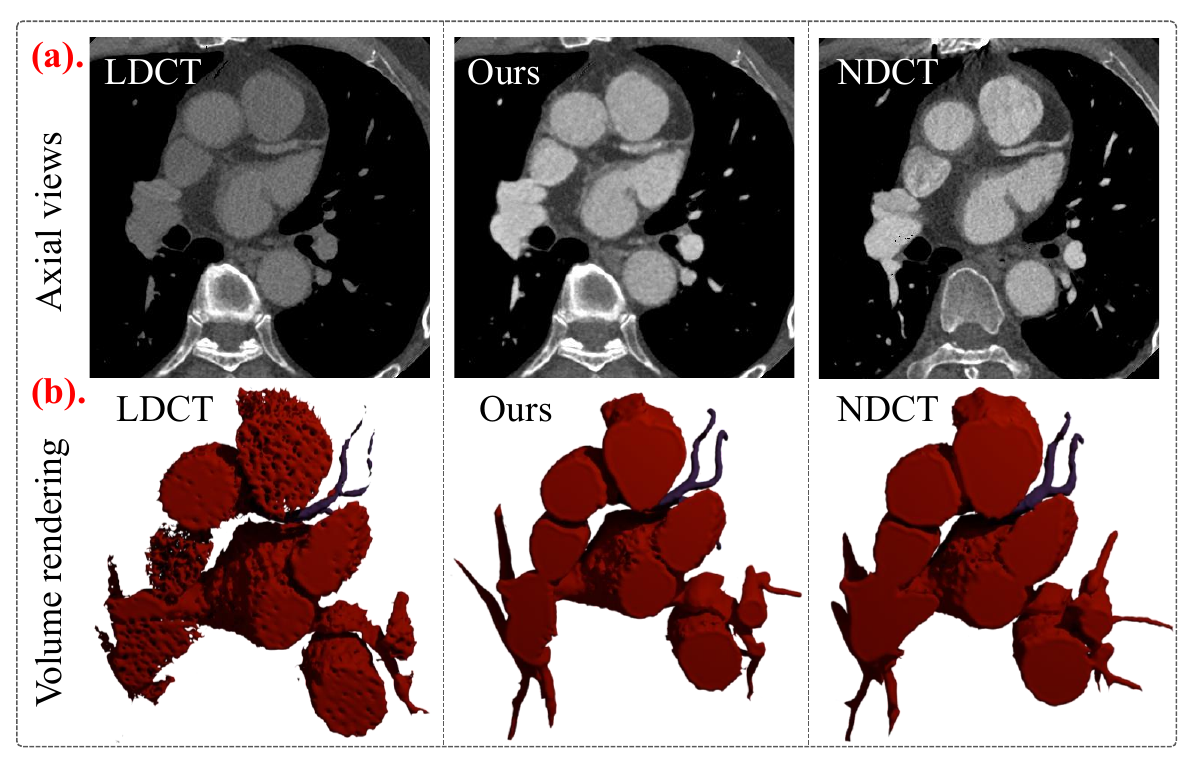}}
\caption{(a) The visualization comparison between our method and LDCT and NDCT from axial views; (b) The coronary volume rendering (VR) image of LDCT, Ours and NDCT.}
\label{fig4}

\end{figure}
\begin{figure*}[htbp]

\centerline{\includegraphics[width=\textwidth]{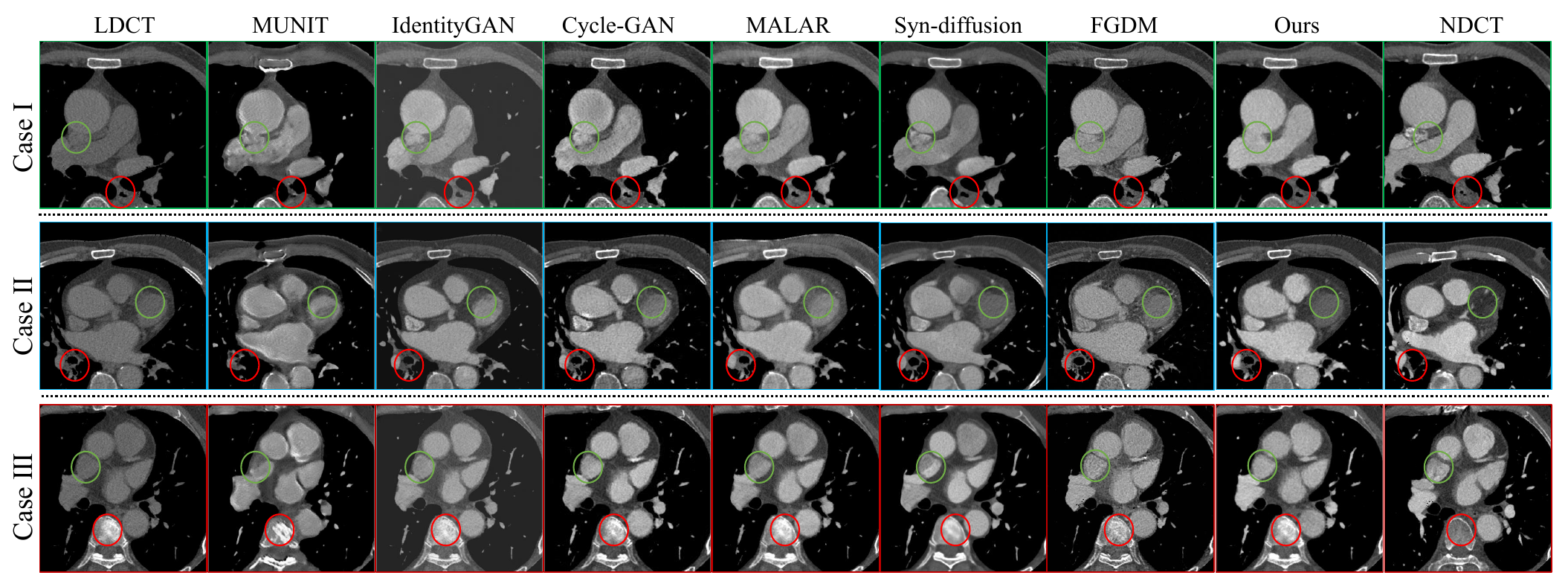}}
\caption{The visualization comparison between our method and state-of-the-art methods. Synthesized images from competing methods are displayed along with LDCT and NDCT (reference) images for representative tasks. The display window for the reconstructed image is set to [-300, 700] HU. The portion of the image in the green and red boxes is the region of interest (ROI). Compared to baselines, proposed method maintains higher anatomical fidelity.}
\label{fig5}

\end{figure*}

\subsection{Experimental Results}
In this section, SLDM qualitative and quantitative results are compared. We illustrate the effectiveness of SLDM in angiographic reconstruction by axial views. Then we evaluate the effectiveness of our SLDM by comparison with state-of-the-art methods and clinical radiographer evaluation, respectively. They are trained and tested on unified LDCT dataset.
\subsubsection{Effectiveness of SLDM in CTA Reconstruction}
Fig. \ref{fig4} shows LDCT, NDCT and reconstruction results. The reconstruction advantages of our proposed SLDM are further demonstrated by comparing with NDCT and LDCT. 
As shown in Fig. \ref{fig4} (a), SLDM can comprehensively enhance narrow coronary arteries with good contrast, smooth vessel interiors, and no distortion in the reconstruction results. 
What's more, the CT intensity of our reconstructed coronary arteries is similar to that of the coronary arteries shown by NDCT and reflects better display than LDCT. 
In addition, the coronary-drawn images in Fig. \ref{fig4} (b) also show the structural integrity of our SLDM results.

\subsubsection{Comparison With the State-of-the-art Methods}

Fig. \ref{fig5} shows the experiments comparing the reconstruction results for three different CT slices.
MUNIT is an unsupervised model and can be observed to exhibit significant structural errors as well as contrast differences in the structures.
IdentityGAN exhibits global shading as well as anomalous enhancements in localized, e.g., skeletal regions, which stems from the fact that it does not have the ability to identify specific structures.
Cycle-GAN, due to its inclusion of the cyclic loss error, the reconstruction results maintain good spatial structure, but its ability to understand content information is poor and the gray values are not uniform.
MALAR shows decent reconstruction of local structure and contrast due to its combination of multiscale local features and remote dependency, but still has undesired contrast deficits in some regions.
Syn-diffusion is diffusion model based medical image translation studies have shown unsatisfactory performance when they are used with precise reconstruction requirements for both structural and contrast information. 
In comparison, FGDM has more consistent structure and gray scale values. 
Unfortunately, its internal structure has more pronounced noise, which can interfere with physician diagnosis.
SLDM addresses these limitations through synergistic integration of topological constraints and multi-semantic guidance, achieving superior boundary accuracy and contrast fidelity. 
The subtraction angiography enhancement module further enables clinically appropriate intensity calibration within optimal HU ranges, enhancing both quantitative performance and diagnostic utility.

The quantitative results in the Table \ref{table 1} show that our method outperforms the comparison algorithms in both reference image evaluation metrics and non-reference image evaluation metrics. It is worth noting that although our method is quite similar to the IdentityGAN method in terms of quantitative assessment, visual assessment and subsequent evaluation by radiologists can further demonstrate the significant advantages of our proposed method over it. In addition, we also conducted a computational complexity analysis on the diffusion model-based method, and the specific analysis results are shown in Appendix B.

\subsubsection{Quantitative Blind Evaluation Results}
We present both the aggregated average scores from the blinded evaluation and the individual average scores assigned by each radiologist in Fig. 6. 
Notably, two of the radiologists consistently assigned higher scores to SLDM results. 
Statistical analysis confirms that SLDM demonstrates statistically significant superiority over other methods across all tested samples.
While GAN-based MALAR and diffusion model-based Syn-diffusion achieved favorable outcomes through certain optimizations, their generalizability and robustness require careful consideration. 
SLDM, by contrast, consistently exhibits strong physician evaluation metrics across diverse datasets, underscoring its promising clinical translation potential.

\begin{figure}[htbp]
\centerline{\includegraphics[width=0.47\textwidth]{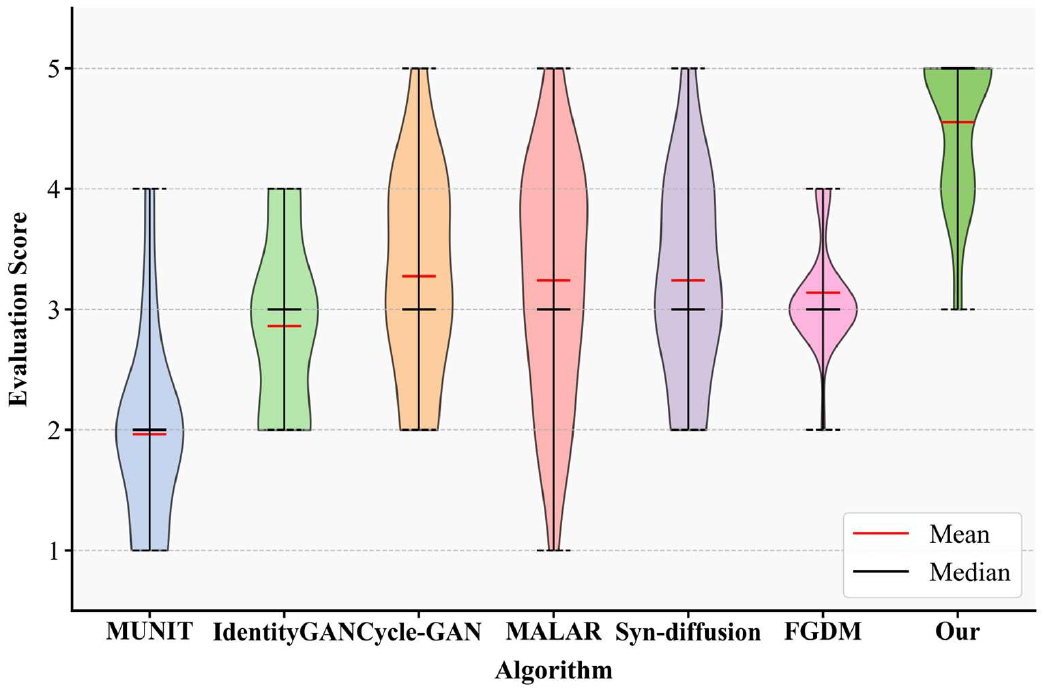}}
\caption{Violin plot analysis of qualitative scores for different methods. Mean scores derived from two radiologists who independently and blindly evaluated each method using real-world LDCT datasets. The red line represents the mean, the black line represents the median, and the area on either side represents the probability density.}
\label{fig6}
\end{figure}

\subsubsection{Segmentation Experiment}
We used the MedSAM \cite{ma2024segment} method to obtain the segmentation results. Fig. 7 shows the segmentation results of the vascular region through different reconstruction methods.
As indicated by the yellow arrows pointing to the target vascular areas, the MUNIT exhibits blurred vessel edges, structural gaps, and background noise artifacts. 
Cycle-GAN exhibits less sharp vessel edges with segmentation discontinuities at tissue junctions. 
IdentityGAN exhibits significant vascular structure errors.
In contrast, the vascular morphology produced by MALAR and Syn-diffusion is smoother and structurally more complete.
However, the consistency of the vascular edge structures is not sufficiently precise. 
FGDM produces jagged vessel edges with unclear boundaries. 
In contrast, our method achieves sharp, continuous vessel with clear boundaries from surrounding tissues. 
It significantly outperforms other methods in preserving fine vascular details, accurately reproducing the morphological and tissue characteristics of vessels in the ground truth.

\begin{figure}[htbp]
\centerline{\includegraphics[width=0.48\textwidth]{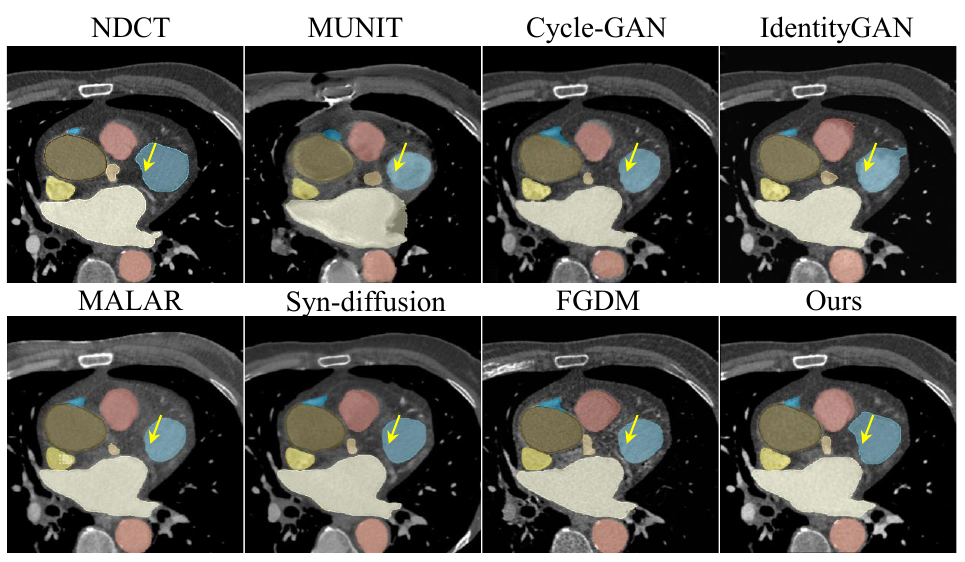}}
\caption{Organ segmentation results between our method and different comparative experiments.}
\label{figseg}
\end{figure}

\subsection{Ablation Study}
To investigate the validity of our proposed method, ablation experiments were carried out on various components and hyperparameters. 
We performed three different ablation experiments to validate the effectiveness of the topology constraint module: without topology Constraint, the topology of single-channel and the topological structure of LDCT.
In addition, the validity of the CTA-CLIP and SAEM were verified. All ablation experiments were conducted on the same dataset.

\subsubsection{Effectiveness of Topology Constraint}
\begin{figure}[htbp]

\centerline{\includegraphics[width=0.45\textwidth]{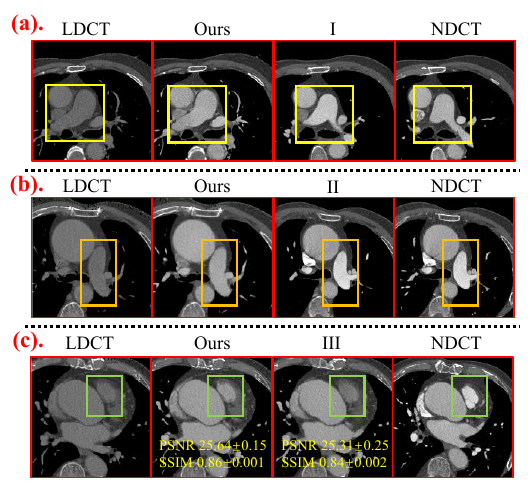}}
\caption{Comparison results from different types of topological datasets. (a) Comparison results from the IR-SDE base framework (I) and and our method; (b) Comparison results obtained with the topological structure of LDCT (II) and our method; (c) Comparison results obtained with the single-channel topology (III) and our method.}
\label{fig7}
\vspace{-0.2cm}
\end{figure}

\begin{figure*}[htbp]
\centerline{\includegraphics[width=\textwidth]{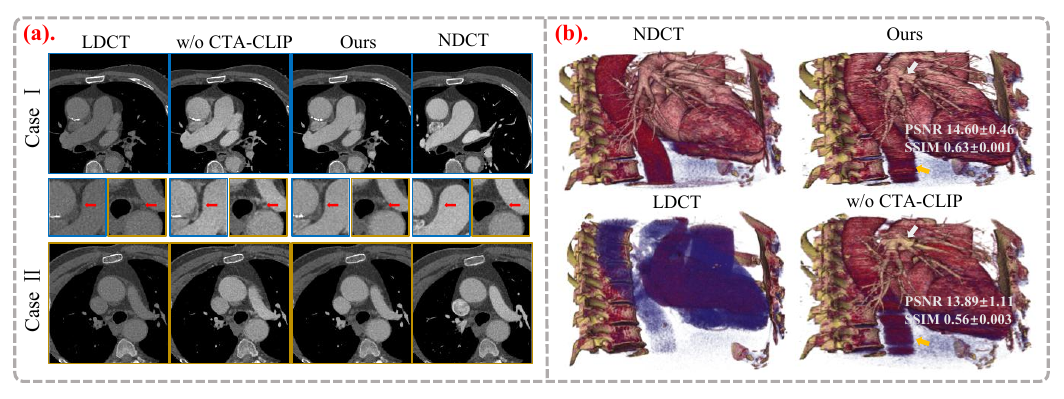}}
\caption{A comparative analysis of reconstruction results with and without CTA-CLIP. (a) illustrates the reconstruction outcomes for various slices, (b) provides a three-dimensional visualization of the overall anatomical structure.}
\label{fig8}
\vspace{-0.2cm}
\end{figure*}

As shown in Fig. \ref{fig7}(a), to systematically evaluate the effectiveness of our topological spatial module, we conducted comparative experiments between the proposed framework and the IR-SDE baseline architecture. 
The results demonstrate that while both approaches achieve contrast enhancement in specific regions confirming effective image domain transformation. 
But the unconstrained IR-SDE method exhibits significant structural aberrations. 
Through analysis, we verified that integrating topological constraints substantially improves structural preservation by maintaining boundary-region connectivity.

  Further validation of our dataset construction strategy was performed through systematic comparison of different image-channel configurations (Fig. \ref{fig7}(b) and Fig. \ref{fig7}(c)).
A critical comparison of topological constraint sources (Fig. \ref{fig7}(b)) demonstrated the limitations of LDCT-derived references. 
It is observed that when the topological space of LDCT is adopted as constraint method, there are error results that do not provide accurate constraint results. 
Therefore, the topological space of NDCT is adopted in our method.
It is important to note that the ablation experimenAts in Fig. 7(a) and 7(b) exhibit results closer to NDCT. 
In fact, it defeats the original purpose of the ICM enhancement algorithm, i.e., the vessel contrast is enhanced but the structure remains unchanged. 
In contrast, our method exhibits more faithful and reasonable results.

  Experimental trials contrasting single-channel grayscale with two-channel contour combination against dual-channel grayscale with single contour configurations revealed superior local structure recovery in the latter paradigm(Fig. \ref{fig7}(c)). 
The objective evaluation indicators presented in Fig. \ref{fig7}(c) show similar agreement.
This empirical evidence supports our hypothesis that reduced cross-domain uncertainty in single-image pair configurations enhances reconstruction fidelity through more deterministic feature mapping.

\subsubsection{Effectiveness of CTA-CLIP}
The effectiveness of the CTA-CLIP was evaluated through comparative analyses of reconstruction performance with and without textual constraints, as demonstrated in Fig. \ref{fig8}.
Comparative analysis reveals that the model without CTA-CLIP  exhibits unintended enhancement during the contrast optimization process (Fig. \ref{fig8} (a)), which may compromise diagnostic reliability.
Integration of text-based semantic supervision effectively regularizes the enhancement trajectory, significantly improving both spatial accuracy and intensity precision in critical anatomical regions.

  The improvement is substantiated through 3D volume rendering analyzes (Fig. \ref{fig8} (b)). 
LDCT reconstructions fail to provide diagnostically meaningful vascular visualization. Our baseline method demonstrates improved vascular delineation capacity at the cost of erroneous enhancement patterns. 
The complete framework incorporating textual guidance achieves vascular enhancement and structural continuity.
The problem of false positives and false negatives that may exist in deep learning network is effectively addressed.
We can also find significant advantages of the text through quantitative indicators.

\begin{figure}[htbp]
\centerline{\includegraphics[width=0.4\textwidth]{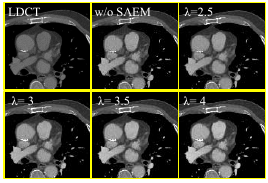}}
\caption{The comparison of results for different parameters $\lambda $ in the SAEM.}
\label{fig9}

\end{figure}

\subsubsection{Effectiveness of SAEM}
\begin{figure}[htbp]
\centerline{\includegraphics[width=0.45\textwidth]{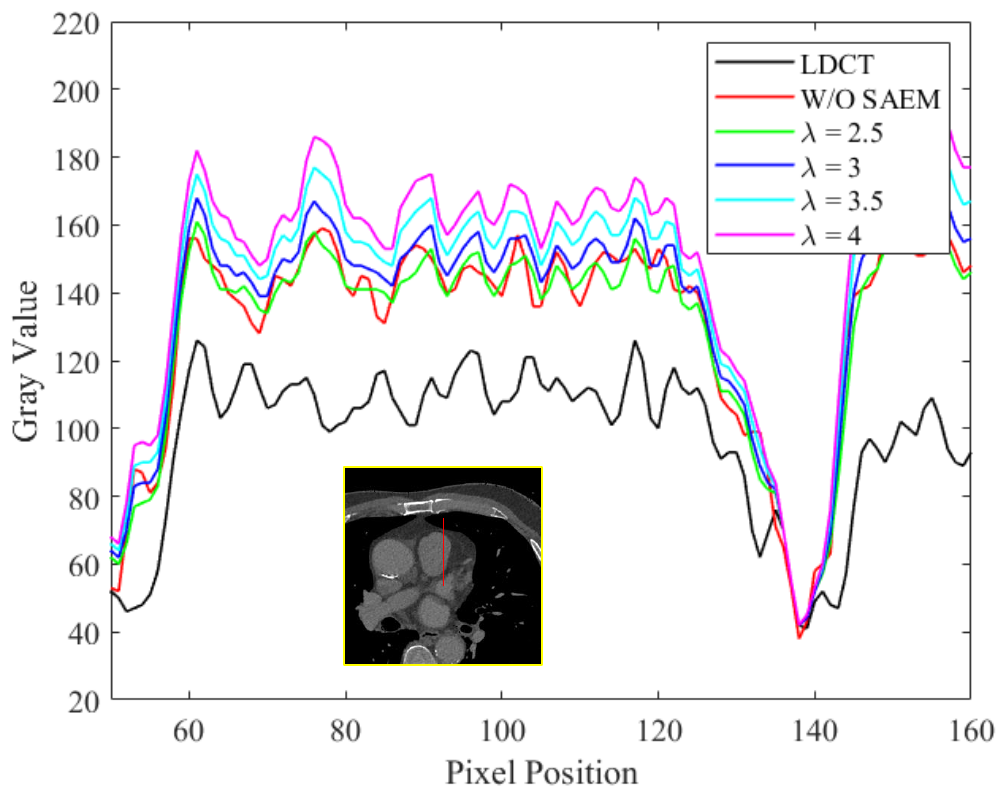}}
\caption{The gray scale histogram of SAEM-free and different weighting parameters $\lambda $.}
\label{fig10}
\end{figure}

The SAEM's performance was evaluated through parametric comparison of SLDM with different weighting configurations, as illustrated in Fig. \ref{fig9}. 
Experimental results demonstrate effective extraction of subtraction angiography features from reconstructed images, enabling precise modulation of contrast ranges in enhanced regions. 
As evidenced in Fig. \ref{fig10}, this functionality allows controlled adjustment of intensity intervals through parameter weighting, significantly improving reconstruction flexibility while maintaining anatomical fidelity. 
The parametric adaptability of our SAEM demonstrates the advantages in balancing contrast optimization with structural preservation in various clinical scenarios.

\section{DISCUSSIONS}
\label{sec:guidelines}

\subsubsection{Uniqueness of the SLDM}
While recent bridge-based diffusion models \cite{liu20232, shi2024resfusion, chung2023direct} demonstrate potential in learning the mapping between distributions, they predominantly necessitate either fully paired datasets or explicit knowledge of the degradation operator. Notably, our proposed method offers a streamlined and effective integration strategy that has the potential to extends these frameworks to the regime of weakly paired data.

Parallel to these developments, conditional generative architectures such as ControlNet \cite{zhang2023adding} and OminiControl \cite{tan2025ominicontrol} leverage structural priors (e.g., edges) to guide image synthesis. However, adapting these methods to the current task presents two fundamental challenges. First, the lack of pixel-wise alignment between LDCT and NDCT data during fine-tuning precludes the use of the LDCT intensity distribution as a data-fidelity constraint. Consequently, the inference process relies exclusively on the extracted LDCT contours. Given that low-dose contours are inherently susceptible to noise artifacts, this unconstrained reliance risks error augmentation. Second, these paradigms typically depend on large-scale foundational models (e.g., Stable Diffusion) adapted via complex fine-tuning procedures, introducing significant computational overhead. In contrast, our approach provides a lightweight, theoretically grounded, and data-efficient solution tailored for real-world medical deployment.

DA-CLIP leverages large-scale pre-trained vision-language models as a universal framework for image restoration \cite{luo2023controlling}. 
Inspired by this approach, All-Inclusive Image Enhancement (AIIE) integrates knowledge of various degradation types into its model, enabling it to handle multiple image degradation tasks \cite{10685536}. 
Dual-Domain CLIP-Assisted Residual Optimization Perception Model (DuDoCROP) fine-tunes DuDoCLIP via contrastive learning in both image and sinogram domains \cite{zhang2024dual}. 
This process extracts semantic descriptions to guide diffusion model optimization. Although these studies provide valuable guidance, CTA reconstruction is a unique task which focuses specifically on enhancing specific structures (like coronary artery).
SLDM innovatively acquires medical categories of key features and encodes them as text embeddings. 
This effectively guides the enhancement of specific structures, offering a novel approach for CTA reconstruction.

\subsubsection{Clinical Benefits of SLDM}
Excessive use of iodinated contrast media (ICM) can cause allergic reactions and acute kidney injury.
This problem is particularly prominent in high-risk patients, such as those with kidney disease and the elderly. 
The SLDM can directly generate enhanced images with equivalent normal-dose quality from low-dose ICM images.
This provides a safer alternative for CTA examinations. 
Through precise structural constraints and text prompts, the SLDM specifically addresses structural distortions and abnormal enhancements in enhancement processes.
This effectively avoids misdiagnosis and missed diagnosis in clinical applications.
In addition, SAEM provides doctors with a personalized optimization scheme.
For special scenarios such as acute aortic dissection, it can help doctors locate the rupture site in a short time by quickly adjusting contrast information.
In the future, integrating SLDM into clinical imaging systems is expected to become a safe and efficient tool.

\subsubsection{Limitations of the SLDM}
Although SLDM has advanced low-dose CTA reconstruction, there are three key limitations that still warrant investigation.
This section delineates each limitation and proposes prospective research directions.
(i) Dependence on weakly paired datasets: 
Although SLDM can perform well with weakly paired datasets, it still causes data collection to become complicated.
Future work should develop fully unsupervised models that utilize a prior information and sampling constraints.
(ii) Lack of real-time capability: On a CT image of 384×384 pixel, SLDM requires about 12 seconds, outperforming conventional diffusion models yet remaining slower than GAN-based approaches.
Achieving real-time reconstruction will likely necessitate integrating consistency model or other single-step generative paradigms.
(iii) Non-contrast CT enhancement: Direct synthesis of NDCT images from non-contrast acquisitions obviating iodinated contrast media holds considerable clinical potential. 
This is a direction we need to further research.

\vspace{-0.3cm}
\section{CONCLUSION}
In conclusion, we proposed a CTA reconstruction method-based diffusion model for enhancing contrast in low-dose ICM.
SLDM is generated using topological space constraints across domains, which effectively maintains structural consistency.
Then, semantic supervision mechanism is introduced to effectively solve the challenge of local structural anomaly enhancement.
The subtraction angiography enhancement module is designed to adjust contrast.
SLDM effectively establishes an accurate conversion from LDCT to NDCT, and the convincing results obtained for result can demonstrate the feasibility and effectiveness of our SLDM.
\vspace{-0.3cm}
\section{APPENDIX}
\subsection{The Proof of Consistency of generated results}
During the training phase, the network learns the conditional score function $\nabla_{\bf{s}} \log p_t({\bf{s}}(t) \mid {\bf{s}}_{\text{gt}}) = -\frac{{\bf{s}}(t)-{\bf{s}}_{\text{gt}}}{v_t}$, where $v_t = \lambda^2(1 - e^{-2\bar{\theta}_t})$ \cite{luo2023image}. During the testing phase, the input structure is ${\bf{s}}_{\text{lq}}$. Assuming the network possesses generalization capabilities, for any conditional structure ${\bf{s}}_{\text{cond}}$, the predicted score is approximated as $-\frac{{\bf{s}}(t)-{\bf{s}}_{\text{cond}}}{v_t}$. Therefore, the drift term for the structural channel in the reverse SDE is given by:
\begin{equation} \frac{d{\bf{s}}}{dt} = \theta_t({\bf{s}}_{\text{lq}} - {\bf{s}}) - \sigma_t^2 \left( -\frac{{\bf{s}} - {\bf{s}}_{\text{lq}}}{v_t} \right) = -\left( \theta_t + \frac{\sigma_t^2}{v_t} \right) ({\bf{s}} - {\bf{s}}_{\text{lq}}). \end{equation}
This constitutes an Ornstein-Uhlenbeck process \cite{maller2009ornstein} with an equilibrium point at ${\bf{s}}_{\text{lq}}$. As $T \to \infty$ and $\int_0^T (\theta_t + \sigma_t^2/v_t)dt \to \infty$, we have $\mathbb{E}[{\bf{s}}(0)] \to {\bf{s}}_{\text{lq}}$ and the variance approaches zero. Thus:
\begin{equation} \lim_{T \to \infty} \mathbb{E}[||{\bf{s}}_0 - {\bf{s}}_{\text{lq}}||_2] = 0. \end{equation}

Let $G$ denote the image generation function of the network, with input $[y_{\text{test}}, {\bf{s}}_{\text{lq}}]$ and output $\hat{x}_0$. Define $G_s([y, s]) = C(G([y, s]))$ as the structure of the generated image. During training, for a training sample $(y, s_{\text{gt}})$, we have $G_s([y, s_{\text{gt}}]) = s_{\text{gt}}$ (perfect alignment). Assume that $G_s$ is Lipschitz continuous, meaning there exists $L_G > 0$ such that:
\begin{equation} |G_s([y, s_1]) - G_s([y, s_2])|_2 \leq L_G |s_1 - s_2|_2. \end{equation}
During testing, the input structure is $s_{\text{lq}}$, and the structure of the generated image is $C(\hat{x}_0) = G_s([y_{\text{test}}, s_{\text{lq}}])$. Consider the following decomposition:
\begin{equation} 
\begin{array}{l}
||{C}({{{\bf{\hat x}}}_0}) - {{\bf{s}}_0}{||_2} = ||{{G}_s}([{{\bf{y}}_{test}},{{\bf{s}}_{lq}}]) - {{\bf{s}}_0}{||_2}\\
 \le ||{{G}_s}([{{\bf{y}}_{test}},{{\bf{s}}_{lq}}]) - {{G}_s}([{{\bf{y}}_{test}},{{\bf{s}}_0}]){||_2} \\
 + ||{{G}_s}([{{\bf{y}}_{test}},{{\bf{s}}_0}]) - {{\bf{s}}_0}{||_2}
\end{array}
\end{equation}
The first term is bounded by Lipschitz continuity: $\leq L_G \|{\bf{s}}_{\text{lq}} - {\bf{s}}_0\|_2$. The second term represents the alignment error: training requires that when the input structure is $s_0$, the generated image structure should be $s_0$. Thus, $\|G_s([y_{\text{test}}, {\bf{s}}_0]) - {\bf{s}}_0\|_2 \leq \epsilon_{\text{align}}$, where $\epsilon_{\text{align}}$ is the alignment error during training. Therefore:
\begin{equation}||C(\hat{x}_0) - s_0||_2 \leq L_G ||{\bf{s}}_0 - {\bf{s}}_{\text{lq}}||_2 + \epsilon_{\text{align}}.\end{equation}
Now, combining this with the triangle inequality:
\begin{equation}
\begin{array}{l}
||C(\hat{x}_0) - {\bf{s}}_{\text{lq}}||_2 \leq \\
||C(\hat{x}_0) - {\bf{s}}_0||_2 + ||{\bf{s}}_0 - {\bf{s}}_{\text{LQ}}||_2 \leq \\
(1 + L_G)||{\bf{s}}_0 - {\bf{s}}_{\text{LQ}}|_2 + \epsilon_{\text{align}}.
\end{array}
\end{equation}
Taking the expectation and letting $K = 1 + L_G$, we obtain:
\begin{equation}\mathbb{E}[||C(\hat{x}_0) - {\bf{s}}_{\text{LQ}}||_2] \leq K \mathbb{E}[||{\bf{s}}_0 - {\bf{s}}_{\text{lq}}|_2] + \epsilon_{\text{align}}.\end{equation}
As $T \to \infty$, $\mathbb{E}[\|{\bf{s}}_0 - {\bf{s}}_{\text{LQ}}\|_2] \to 0$, hence $\mathbb{E}[\|C(\hat{x}_0) - {\bf{s}}_{\text{LQ}}\|_2] \leq \epsilon_{\text{align}}$. If training is perfect ($\epsilon_{\text{align}} = 0$), then $C(\hat{x}_0)$ is exactly equal to ${\bf{s}}_{\text{LQ}}$. In the general case, the error is controlled by $K \mathbb{E}[\|{\bf{s}}_0 - {\bf{s}}_{\text{LQ}}\|_2]$.

\subsection{Computational complexity of the method}

To comprehensively evaluate the practical deployment efficiency of each diffusion-based method, we conducted a statistical analysis of the inference time and model parameter count of diffusion-based approaches. The results are shown in Table \ref{table2}.
In terms of inference time, our proposed method requires only 7.2 seconds, achieving a 57× speedup compared to FGDM (410.2 seconds) while maintaining superior generation quality. This significant improvement is primarily attributed to our optimized diffusion sampling strategy. Although Syndiff achieves the fastest inference speed (0.4 seconds), its generation quality is relatively limited and falls short of practical application requirements.
Regarding model parameter count, our method totals 295.2M parameters (49.0M + 246.2M), with the CLIP module accounting for the majority of parameters and providing robust semantic understanding capabilities. In comparison, FGDM's bidirectional diffusion architecture has a total of 57.2M parameters, while Syndiff contains 61.4M parameters. 
Overall, our method achieves a better balance between inference efficiency and model performance, ensuring high-quality generation results while remaining feasible for practical deployment.

\begin{table}[htbp]
\centering
\caption{Statistical results of inference time and model parameters for diffusion-based methods}
\label{table2}
\begin{tabular}{cccc} 
\toprule
Method & Inference Time (s) & \multicolumn{2}{c}{Model Parameters (Million)} \\ 
\midrule
FGDM & 410.2 & \multicolumn{2}{c}{61.4} \\ 
\hline
\multirow{2}{*}{Syndiff} & \multirow{2}{*}{0.4} & DiffusionAtoB & 28.6 \\ 
\cmidrule{3-4}
 & & DiffusionBtoA & 28.6 \\ 
\hline
\multirow{2}{*}{Ours} & \multirow{2}{*}{7.2} & Diffusion\_Module & 49.0 \\ 
\cmidrule{3-4}
 & & CLIP\_module & 246.2 \\
\midrule
Total Parameters & - & FGDM & 61.4 \\
 & & Syndiff & 57.2 \\
 & & Ours & 295.2 \\
\bottomrule
\end{tabular}
\end{table}

\bibliographystyle{IEEEtran}

\bibliography{ref}

@article{fahling2017understanding,
  title={Understanding and preventing contrast-induced acute kidney injury},
  author={F{\"a}hling, Michael and Seeliger, Erdmann and Patzak, Andreas and Persson, Pontus B},
  journal={Nature Reviews Nephrology},
  volume={13},
  number={3},
  pages={169--180},
  year={2017},
  publisher={Nature Publishing Group UK London}
}

@article{yu2007review,
  title={Review of CT angiography of aorta},
  author={Yu, Tongfu and Zhu, Xiaomei and Tang, Lijun and Wang, Dehang and Saad, Nael},
  journal={Radiologic clinics of North america},
  volume={45},
  number={3},
  pages={461--483},
  year={2007},
  publisher={Elsevier}
}

@article{chiu2022hypersensitivity,
  title={Hypersensitivity reactions to iodinated contrast media},
  author={Chiu, Tsu-Man and Chu, Sung-Yu},
  journal={Biomedicines},
  volume={10},
  number={5},
  pages={1036},
  year={2022},
  publisher={MDPI}
}

@article{rubin1997helical,
  title={Helical CT angiography of the thoracic aorta},
  author={Rubin, Geoffrey D},
  journal={Journal of thoracic imaging},
  volume={12},
  number={2},
  pages={128--149},
  year={1997},
  publisher={LWW}
}

@article{golledge2008acute,
  title={Acute aortic dissection},
  author={Golledge, Jonathan and Eagle, Kim A},
  journal={The Lancet},
  volume={372},
  number={9632},
  pages={55--66},
  year={2008},
  publisher={Elsevier}
}

@inproceedings{10.1117/12.2581056,
author = {Huiqiao Xie and Yang Lei and Tonghe Wang and Pretesh Patel and Walter J. Curran and Tian Liu and Xiangyang Tang and Xiaofeng Yang},
title = {Generation of contrast-enhanced CT with residual cycle-consistent generative adversarial network (Res-CycleGAN)},
volume = {11595},
booktitle = {Medical Imaging 2021: Physics of Medical Imaging},
editor = {Hilde Bosmans and Wei Zhao and Lifeng Yu},
organization = {International Society for Optics and Photonics},
publisher = {SPIE},
pages = {1159540},
year = {2021},
doi = {10.1117/12.2581056},
URL = {https://doi.org/10.1117/12.2581056}
}

@article{kim2021feasibility,
  title={The feasibility of deep learning-based synthetic contrast-enhanced CT from nonenhanced CT in emergency department patients with acute abdominal pain},
  author={Kim, Se Woo and Kim, Jung Hoon and Kwak, Suha and Seo, Minkyo and Ryoo, Changhyun and Shin, Cheong-Il and Jang, Siwon and Cho, Jungheum and Kim, Young-Hoon and Jeon, Kyutae},
  journal={Scientific reports},
  volume={11},
  number={1},
  pages={20390},
  year={2021},
  publisher={Nature Publishing Group UK London}
}

@article{mccollough2015dual,
  title={Dual-and multi-energy CT: principles, technical approaches, and clinical applications},
  author={McCollough, Cynthia H and Leng, Shuai and Yu, Lifeng and Fletcher, Joel G},
  journal={Radiology},
  volume={276},
  number={3},
  pages={637--653},
  year={2015},
  publisher={Radiological Society of North America}
}

@article{zhang2022multiple,
  title={Multiple adversarial learning based angiography reconstruction for ultra-low-dose contrast medium CT},
  author={Zhang, Weiwei and Zhou, Zhen and Gao, Zhifan and Yang, Guang and Xu, Lei and Wu, Weiwen and Zhang, Heye},
  journal={IEEE Journal of Biomedical and Health Informatics},
  volume={27},
  number={1},
  pages={409--420},
  year={2022},
  publisher={IEEE}
}

@article{gleeson2004contrast,
  title={Contrast-induced nephropathy},
  author={Gleeson, Tadhg G and Bulugahapitiya, Sudi},
  journal={American Journal of Roentgenology},
  volume={183},
  number={6},
  pages={1673--1689},
  year={2004},
  publisher={American Roentgen Ray Society}
}

@article{kang2019cycle,
  title={Cycle-consistent adversarial denoising network for multiphase coronary CT angiography},
  author={Kang, Eunhee and Koo, Hyun Jung and Yang, Dong Hyun and Seo, Joon Bum and Ye, Jong Chul},
  journal={Medical physics},
  volume={46},
  number={2},
  pages={550--562},
  year={2019},
  publisher={Wiley Online Library}
}

@article{lyu2023generative,
  title={Generative adversarial network-based non-contrast CT angiography for aorta and carotid arteries},
  author={Lyu, Jinhao and Fu, Ying and Yang, Mingliang and Xiong, Yongqin and Duan, Qi and Duan, Caohui and Wang, Xueyang and Xing, Xinbo and Zhang, Dong and Lin, Jiaji and others},
  journal={Radiology},
  volume={309},
  number={2},
  pages={e230681},
  year={2023},
  publisher={Radiological Society of North America}
}

@inproceedings{zhu2017unpaired,
  title={Unpaired image-to-image translation using cycle-consistent adversarial networks},
  author={Zhu, Jun-Yan and Park, Taesung and Isola, Phillip and Efros, Alexei A},
  booktitle={Proceedings of the IEEE international conference on computer vision},
  pages={2223--2232},
  year={2017}
}

@article{pang2023ncct,
  title={NCCT-CECT image synthesizers and their application to pulmonary vessel segmentation},
  author={Pang, Haowen and Qi, Shouliang and Wu, Yanan and Wang, Meihuan and Li, Chen and Sun, Yu and Qian, Wei and Tang, Guoyan and Xu, Jiaxuan and Liang, Zhenyu and others},
  journal={Computer Methods and Programs in Biomedicine},
  volume={231},
  pages={107389},
  year={2023},
  publisher={Elsevier}
}

@article{ozbey2023unsupervised,
  title={Unsupervised medical image translation with adversarial diffusion models},
  author={{\"O}zbey, Muzaffer and Dalmaz, Onat and Dar, Salman UH and Bedel, Hasan A and {\"O}zturk, {\c{S}}aban and G{\"u}ng{\"o}r, Alper and {\c{C}}ukur, Tolga},
  journal={IEEE Transactions on Medical Imaging},
  volume={42},
  number={12},
  pages={3524--3539},
  year={2023},
  publisher={IEEE}
}

@article{dhariwal2021diffusion,
  title={Diffusion models beat gans on image synthesis},
  author={Dhariwal, Prafulla and Nichol, Alexander},
  journal={Advances in neural information processing systems},
  volume={34},
  pages={8780--8794},
  year={2021}
}

@article{wu2023wavelet,
  title={Wavelet-improved score-based generative model for medical imaging},
  author={Wu, Weiwen and Wang, Yanyang and Liu, Qiegen and Wang, Ge and Zhang, Jianjia},
  journal={IEEE transactions on medical imaging},
  volume={43},
  number={3},
  pages={966--979},
  year={2023},
  publisher={IEEE}
}

@article{wu2024multi,
  title={Multi-channel optimization generative model for stable ultra-sparse-view CT reconstruction},
  author={Wu, Weiwen and Pan, Jiayi and Wang, Yanyang and Wang, Shaoyu and Zhang, Jianjia},
  journal={IEEE Transactions on Medical Imaging},
  year={2024},
  publisher={IEEE}
}

@inproceedings{huang2018multimodal,
  title={Multimodal unsupervised image-to-image translation},
  author={Huang, Xun and Liu, Ming-Yu and Belongie, Serge and Kautz, Jan},
  booktitle={Proceedings of the European conference on computer vision (ECCV)},
  pages={172--189},
  year={2018}
}

@article{luo2023controlling,
  title={Controlling vision-language models for universal image restoration},
  author={Luo, Ziwei and Gustafsson, Fredrik K and Zhao, Zheng and Sj{\"o}lund, Jens and Sch{\"o}n, Thomas B},
  journal={arXiv preprint arXiv:2310.01018},
  volume={3},
  number={8},
  year={2023}
}

@inproceedings{luo2023image,
  title={Image Restoration with Mean-Reverting Stochastic Differential Equations},
  author={Luo, Ziwei and Gustafsson, Fredrik K and Zhao, Zheng and Sj{\"o}lund, Jens},
  booktitle={International Conference on Machine Learning (ICML), Honolulu, Hawaii, USA, 23-29 July, 2023},
  volume={202},
  pages={23045--23066},
  year={2023}
}

@inproceedings{zhang2022contrastive,
  title={Contrastive learning of medical visual representations from paired images and text},
  author={Zhang, Yuhao and Jiang, Hang and Miura, Yasuhide and Manning, Christopher D and Langlotz, Curtis P},
  booktitle={Machine learning for healthcare conference},
  pages={2--25},
  year={2022},
  organization={PMLR}
}

@inproceedings{chen2024low,
  title={Low-dose CT denoising with language-engaged dual-space alignment},
  author={Chen, Zhihao and Chen, Tao and Wang, Chenhui and Gao, Qi and Niu, Chuang and Wang, Ge and Shan, Hongming},
  booktitle={2024 IEEE International Conference on Bioinformatics and Biomedicine (BIBM)},
  pages={3088--3091},
  year={2024},
  organization={IEEE}
}

@article{dalmaz2022resvit,
  title={ResViT: residual vision transformers for multimodal medical image synthesis},
  author={Dalmaz, Onat and Yurt, Mahmut and {\c{C}}ukur, Tolga},
  journal={IEEE Transactions on Medical Imaging},
  volume={41},
  number={10},
  pages={2598--2614},
  year={2022},
  publisher={IEEE}
}

@inproceedings{radford2021learning,
  title={Learning transferable visual models from natural language supervision},
  author={Radford, Alec and Kim, Jong Wook and Hallacy, Chris and Ramesh, Aditya and Goh, Gabriel and Agarwal, Sandhini and Sastry, Girish and Askell, Amanda and Mishkin, Pamela and Clark, Jack and others},
  booktitle={International conference on machine learning},
  pages={8748--8763},
  year={2021},
  organization={PmLR}
}

@article{mokady2021clipcap,
  title={Clipcap: Clip prefix for image captioning},
  author={Mokady, Ron and Hertz, Amir and Bermano, Amit H},
  journal={arXiv preprint arXiv:2111.09734},
  year={2021}
}

@inproceedings{rombach2022high,
  title={High-resolution image synthesis with latent diffusion models},
  author={Rombach, Robin and Blattmann, Andreas and Lorenz, Dominik and Esser, Patrick and Ommer, Bj{\"o}rn},
  booktitle={Proceedings of the IEEE/CVF conference on computer vision and pattern recognition},
  pages={10684--10695},
  year={2022}
}

@article{niu2023ct,
  title={CT multi-task learning with a Large Image-Text (LIT) model},
  author={Niu, Chuang and Wang, Ge},
  journal={bioRxiv},
  pages={2023--04},
  year={2023},
  publisher={Cold Spring Harbor Laboratory}
}

@article{li2023llava,
  title={Llava-med: Training a large language-and-vision assistant for biomedicine in one day},
  author={Li, Chunyuan and Wong, Cliff and Zhang, Sheng and Usuyama, Naoto and Liu, Haotian and Yang, Jianwei and Naumann, Tristan and Poon, Hoifung and Gao, Jianfeng},
  journal={Advances in Neural Information Processing Systems},
  volume={36},
  pages={28541--28564},
  year={2023}
}

@article{10685536,
  title={All-Inclusive image enhancement for degraded images exhibiting low-frequency corruption},
  journal={IEEE Transactions on Circuits and Systems for Video Technology}, 
  author={Ju, Mingye and He, Chunming and Ding, Can and Ren, Wenqi and Zhang, Lin and Ma, Kai-Kuang},
  year={2025},
  volume={35},
  number={1},
  pages={838-856},
  doi={10.1109/TCSVT.2024.3465875}
}

@article{zhang2024dual,
  title={Dual-domain CLIP-assisted residual optimization perception model for metal artifact reduction},
  author={Zhang, Xinrui and Cai, Ailong and Wang, Shaoyu and Wang, Linyuan and Zheng, Zhizhong and Li, Lei and Yan, Bin},
  journal={arXiv preprint arXiv:2408.14342},
  year={2024}
}

@article{jin2024llmra,
  title={Llmra: Multi-modal large language model based restoration assistant},
  author={Jin, Xiaoyu and Shi, Yuan and Xia, Bin and Yang, Wenming},
  journal={arXiv preprint arXiv:2401.11401},
  year={2024}
}

@ARTICLE{10287612,
  author={Li, Yunxiang and Shao, Hua-Chieh and Liang, Xiao and Chen, Liyuan and Li, Ruiqi and Jiang, Steve and Wang, Jing and Zhang, You},
  journal={IEEE Transactions on Medical Imaging}, 
  title={Zero-Shot Medical Image Translation via Frequency-Guided Diffusion Models}, 
  year={2024},
  volume={43},
  number={3},
  pages={980-993},
  doi={10.1109/TMI.2023.3325703}}

@article{gonzalez2018image,
  title={Image-to-image translation for cross-domain disentanglement},
  author={Gonzalez-Garcia, Abel and Van De Weijer, Joost and Bengio, Yoshua},
  journal={Advances in neural information processing systems},
  volume={31},
  year={2018}
}

@inproceedings{liu2024structure,
  title={Structure matters: Tackling the semantic discrepancy in diffusion models for image inpainting},
  author={Liu, Haipeng and Wang, Yang and Qian, Biao and Wang, Meng and Rui, Yong},
  booktitle={Proceedings of the IEEE/CVF Conference on Computer Vision and Pattern Recognition},
  pages={8038--8047},
  year={2024}
}

@article{arjovsky2017towards,
  title={Towards principled methods for training generative adversarial networks},
  author={Arjovsky, Martin and Bottou, L{\'e}on},
  journal={arXiv preprint arXiv:1701.04862},
  year={2017}
}

@article{yue2023image,
  title={Image restoration through generalized ornstein-uhlenbeck bridge},
  author={Yue, Conghan and Peng, Zhengwei and Ma, Junlong and Du, Shiyan and Wei, Pengxu and Zhang, Dongyu},
  journal={arXiv preprint arXiv:2312.10299},
  year={2023}
}

@article{ma2024segment,
  title={Segment anything in medical images},
  author={Ma, Jun and He, Yuting and Li, Feifei and Han, Lin and You, Chenyu and Wang, Bo},
  journal={Nature Communications},
  volume={15},
  number={1},
  pages={654},
  year={2024},
  publisher={Nature Publishing Group UK London}
}

@inproceedings{chambon2022adapting,
  title={Adapting Pretrained Vision-Language Foundational Models to Medical Imaging Domains},
  author={Chambon, Pierre Joseph Marcel and Bluethgen, Christian and Langlotz, Curtis and Chaudhari, Akshay},
  booktitle={NeurIPS 2022 Foundation Models for Decision Making Workshop},
  year={2024}
}

@inproceedings{liu20232,
  title={I $^2$ SB: Image-to-Image Schr{\"o}dinger Bridge},
  author={Liu, Guan-Horng and Vahdat, Arash and Huang, De-An and Theodorou, Evangelos and Nie, Weili and Anandkumar, Anima},
  booktitle={International Conference on Machine Learning},
  pages={22042--22062},
  year={2023},
  organization={PMLR}
}

@article{shi2024resfusion,
  title={Resfusion: Denoising diffusion probabilistic models for image restoration based on prior residual noise},
  author={Shi, Zhenning and Zheng, Haoshuai and Xu, Chen and Dong, Changsheng and Pan, Bin and Xie, Xueshuo and He, Along and Li, Tao and Fu, Huazhu},
  journal={Advances in Neural Information Processing Systems},
  volume={37},
  pages={130664--130693},
  year={2024}
}

@article{maller2009ornstein,
  title={Ornstein--Uhlenbeck processes and extensions},
  author={Maller, Ross A and M{\"u}ller, Gernot and Szimayer, Alex},
  journal={Handbook of financial time series},
  pages={421--437},
  year={2009},
  publisher={Springer}
}

@article{chung2023direct,
  title={Direct diffusion bridge using data consistency for inverse problems},
  author={Chung, Hyungjin and Kim, Jeongsol and Ye, Jong Chul},
  journal={Advances in Neural Information Processing Systems},
  volume={36},
  pages={7158--7169},
  year={2023}
}

@inproceedings{zhang2023adding,
  title={Adding conditional control to text-to-image diffusion models},
  author={Zhang, Lvmin and Rao, Anyi and Agrawala, Maneesh},
  booktitle={Proceedings of the IEEE/CVF international conference on computer vision},
  pages={3836--3847},
  year={2023}
}

@inproceedings{tan2025ominicontrol,
  title={Ominicontrol: Minimal and universal control for diffusion transformer},
  author={Tan, Zhenxiong and Liu, Songhua and Yang, Xingyi and Xue, Qiaochu and Wang, Xinchao},
  booktitle={Proceedings of the IEEE/CVF International Conference on Computer Vision},
  pages={14940--14950},
  year={2025}
}
\vspace{-0.9cm}

\end{document}